\documentclass{article} 
\usepackage[final]{colm2025_conference}

\usepackage{microtype}
\usepackage{hyperref}
\usepackage{url}
\usepackage{booktabs}

\usepackage{lineno}

\usepackage{graphicx}
\usepackage[whole]{bxcjkjatype}
\usepackage{makecell}
\usepackage{amsfonts} 

\definecolor{darkblue}{rgb}{0, 0, 0.5}
\hypersetup{colorlinks=true, citecolor=darkblue, linkcolor=darkblue, urlcolor=darkblue}

\title{Why We Build Local Large Language Models: An Observational Analysis from 35 Japanese and Multilingual LLMs}

\author{
\textbf{Koshiro Saito}\textsuperscript{1}\quad
\textbf{Sakae Mizuki}\textsuperscript{2,1}\quad
\textbf{Masanari Ohi}\textsuperscript{1}\quad
\textbf{Taishi Nakamura}\textsuperscript{1}\quad \\
\textbf{Taihei Shiotani}\textsuperscript{1}\quad
\textbf{Koki Maeda}\textsuperscript{1}\quad
\textbf{Youmi Ma}\textsuperscript{1}\quad
\textbf{Kakeru Hattori}\textsuperscript{1}\quad
\textbf{Kazuki Fujii}\textsuperscript{1}\quad \\
\textbf{Takumi Okamoto}\textsuperscript{1}\quad
\textbf{Shigeki Ishida}\textsuperscript{1}\quad
\textbf{Hiroya Takamura}\textsuperscript{2}\quad
\textbf{Rio Yokota}\textsuperscript{1}\quad
\textbf{Naoaki Okazaki}\textsuperscript{1,2,3}\\
[0.5em]
\textsuperscript{1}Institute of Science Tokyo \\
\textsuperscript{2}National Institute of Advanced Industrial Science and Technology \\
\textsuperscript{3}NII LLMC \\ \\
\texttt{\{koshiro.saito,sakae.mizuki,ohi.masanari,nakamura.taishi,shiotani.taihei,}\\
\texttt{koki.maeda,ma.youmi\}@nlp.comp.isct.ac.jp, kakeru.hattori@nlp.c.titech.ac.jp,}\\
\texttt{\{taishi,kazuki.fujii,okamoto,ishida,rioyokota\}@rio.scrc.iir.isct.ac.jp,}\\
\texttt{hiroya.takamura@aist.go.jp}
}

\begin{document}

\ifcolmsubmission
\linenumbers
\fi

\maketitle

\begin{abstract}
Why do we build local large language models (LLMs)?
What should a local LLM learn from the target language? Which abilities can be transferred from other languages?
Do language-specific scaling laws exist?
To explore these research questions, we evaluated 35 Japanese, English, and multilingual LLMs on 19 evaluation benchmarks for Japanese and English, taking Japanese as a local language.
Adopting an observational approach, we analyzed correlations of benchmark scores, and conducted principal component analysis (PCA) to derive \textit{ability factors}.
We found that if LLMs perform well in English on tasks like academic subjects, code generation, arithmetic reasoning, commonsense, and reading comprehension, they also perform well on the same tasks in Japanese. 
This indicates it is not necessary to specifically train on Japanese text to enhance abilities for solving these tasks.
In contrast, training on Japanese text improves question-answering tasks about Japanese knowledge and English-Japanese translation, which indicates that abilities for solving these two tasks can be regarded as \textit{Japanese abilities}.
Furthermore, we confirmed that the Japanese abilities scale with the computational budget for Japanese text. Taken together, our findings offer generalizable insights into which tasks benefit from local-language data and what we can expect when building local LLMs.
\end{abstract}

\section{Introduction} \label{intro}
Major large language models (LLMs) are English-centric (\textit{English LLMs} hereafter), e.g., Meta Llama 3~\citep{metallama3}, Mistral~\citep{jiang2023mistral7b}, and Phi-3~\citep{phi_3},
due to the dominance of English on the internet and the global economy, which results in a limited focus on non-English languages.
Several companies and research institutes have been actively developing LLMs targeting non-English languages (\textit{local LLMs} hereafter), e.g., Bllossom~\citep{choi-etal-2024-optimizing}, Chinese-LLaMA~\citep{cui2024efficienteffectivetextencoding} and openCabrita~\citep{larcher2023cabritaclosinggapforeign}, driven by various motivations. These include advancing research and development in multilingual NLP, mitigating security risks associated with relying on a limited number of foreign companies, and promoting responsible artificial intelligence for their community.

However, the advantages of training LLMs on non-English text remain underexplored--particularly regarding the unique skills or knowledge such LLMs might gain compared to English-centric or Multilingual LLMs.
On the one hand, LLMs have demonstrated high multilingual abilities, such as arithmetic reasoning~\citep{shi2023language} and machine translation \citep{briakou-etal-2023-searching}, which casts doubt on the advantage of training on non-English text. On the other hand, training on non-English text has been reported to bring stronger cultural and regional knowledge of the target language~\citep{includeromanou2025}, although there are mixed findings for other tasks such as commonsense reasoning and reading comprehension~\citep{cui2024efficienteffectivetextencoding, choi-etal-2024-optimizing, larcher2023cabritaclosinggapforeign}. These two perspectives--multilinguality versus language specificity--suggest that the effectiveness of training on non-English text is inherently task dependent.
Indeed, demonstrating an advantage of training on non-English text remains not straightforward. Numerious studies have built non-English LLMs from scratch~\citep{holmstrom-etal-2023-bridging} or via continual pre-training (CPT) over English LLMs~\citep{cui2024efficienteffectivetextencoding, choi-etal-2024-optimizing, larcher2023cabritaclosinggapforeign}, but their task-specific results are often mixed or contradictly, raising doubts about generalizability (\S~\ref{subsec:related_work_training_on_non_english_text}). Because LLM performances depends on several design choices--such as training from scratch or via CPT, which base model is selected for CPT~\citep{tejaswi2024exploringdesignchoicesbuilding}, and how the training data is curated~\citep{finewebpenedo2024, datacomplmli2024}--it is difficult to isolate performance gains specifically attributable to training on non-English text.
Given its huge impact, thorough investigation and convincing insights into the advantages of local LLMs are valuable.

To explore what unique skills or knowledge may emerge as the natural consequence of the training on non-English text, we adopt an observational approach~\citep{ruan2024observationalscalinglawspredictability} for Japanese-centric LLMs (\textit{Japanese LLMs} hereafter), leveraging the exceptionally active development in Japan (e.g., Llama 3.1 Swallow\footnote{\url{https://swallow-llm.github.io/llama3-swallow.en.html}} and LLM-jp~\citep{llmjp2024llmjpcrossorganizationalprojectresearch}) among non-English initiatives. Specifically, we evaluate 35 publicly available Japanese, English, and multilingual LLMs representing a variety of design choices. We also use 19 comprehensive evaluation benchmarks covering knowledge-based QA, academic subjects, reading comprehension, and more, tasked in Japanese and English. These also includes paired Japanese and English benchmarks so that we can compare the task performance across both languages. Our goal is to derive generalizable insights (i.e., insights that are robust to design choices) by conducting a quantitative analysis.

First, to explore multilinguality versus language specificity, we analyzed score correlations across 19 task benchmarks for 35 LLMs, and applied Principal Component Analysis (PCA) to represent the performance in a low-dimensional \textit{ability space}~\citep{ruan2024observationalscalinglawspredictability}. We found that tasks such as academic subjects, code generation, and arithmetic reasoning exhibited strong cross-lingual correlations on their scores and were associated with the same ability factors across languages. This indicates strong multilingual transferability, suggesting that training in English text would also improve performance on these tasks when tested in Japanese. Conversely, tasks such as QA about Japanese cultural knowledge and English-Japanese translation exhibited weak correlations with other tasks and were strongly associated with an independent ability factor, indicating language-specific abilities.

Second, to investigate the language-specific abilities attributed to training on Japanese text, we examined language-specific scaling laws. Specifically, we defined the language-specific computational budget as the product of the number of parameters and training tokens for each language~\citep{chinchilla-scaling-laws}, and analyzed the log-linear relationship between these budgets and the ability factors obtained by PCA. We found that the English computational budget showed a strong correlation with the general ability factor but a weak correlation with the Japanese-specific ability factor. In contrast, the Japanese computational budget showed a strong correlation with the Japanese ability factor, suggesting that enhancement of Japanese knowledge and English-Japanese translation skills arise from training on Japanese text itself beyond particular design choice. These knowledge and skill scale with the amount of Japanese training text and are difficult to acquire solely from English text.

\section{Related Work}
\label{sec:related_work}

\subsection{Effects of Training on Non-English Text}
\label{subsec:related_work_training_on_non_english_text}

There is a growing number of studies examining the impacts of training local LLMs on target language data:
Chinese~\citep{zhao2024llamaenglishempiricalstudy, cui2024efficienteffectivetextencoding},
Turkish~\citep{toraman-2024-adapting},
Portuguese~\citep{larcher2023cabritaclosinggapforeign},
Swedish~\citep{holmstrom-etal-2023-bridging},
and Finnish~\citep{luukkonen-etal-2023-fingpt}.
Some studies consistently reported gains in reading comprehension~\citep{etxaniz-etal-2024-latxa, fujii2024continual, dou-etal-2024-sailor, joshi-etal-2025-adapting, vo2024redwhaleadaptedkoreanllm, larcher2023cabritaclosinggapforeign}, commonsense reasoning~\citep{etxaniz-etal-2024-latxa, fujii2024continual, 10799288, dou-etal-2024-sailor, joshi-etal-2025-adapting, vo2024redwhaleadaptedkoreanllm, choi-etal-2024-optimizing, tejaswi2024exploringdesignchoicesbuilding-arxiv}, and local knowledge QA~\citep{etxaniz-etal-2024-latxa, fujii2024continual, joshi-etal-2025-adapting, NEURIPS2024_3bb42f6b}.
However, following our survey of 15 previous reports on non-English LLMs (see Table~\ref{tab:prior_works} in \S~\ref{app:comp_w_prior}), the evidence remains fragmented for two reasons:
1) Sparse coverage of task types:
Prior works evaluated only a small set of benchmarks (an average of 2.5). In particular, machine-translation and coding tasks appear in just 2 and 0 out of 15 studies, respectively.
2) Contradictory results: Some studies drew (self-)contradictory conclusions: e.g., for mathematical reasoning, \citet{etxaniz-etal-2024-latxa} reported positive+neutral effects, whereas \citet{pipatanakul2023typhoonthailargelanguage} noted negative+neutral effects; for academic subject, both of \citet{10799288} and \citet{dou-etal-2024-sailor} documented positive+neutral effects; and, for summarization, \citet{fujii2024continual} observed a negative effect, whereas \citet{joshi-etal-2025-adapting} and \citet{tejaswi2024exploringdesignchoicesbuilding-arxiv} found a positive effect. 

\subsection{Multilinguality vs Language-specificity}
Training on non-English corpora sometimes involve using multilingual corpora. 
\citet{berend-2022-combating} and \citet{chang-etal-2024-multilinguality} reported that multilingual training does not always improve performance due to the curse of multilingualty~\citep{conneau-etal-2020-unsupervised}.
Furthermore, English and multilingual LLMs reportedly show strong multilingual abilities on tasks such as arithmetic and commonsense reasoning~\citep{shi2023language} through cross-language generalization~\citep{zhang-etal-2023-dont}. These findings suggest that the benefits of training on non-English text might be limited or task-dependent.

\subsection{Correlations between Tasks and Ability Factors}
Several prior studies have investigated the correlations between different task benchmarks and associated the task performance with a small number of ability factors~\citep{ruan2024observationalscalinglawspredictability, ni2024mixevalderivingwisdomcrowd, tiong-etal-2024-measuring}. These studies have reported strong correlations between knowledge-based QA tasks and identified ability factors specific to arithmetic reasoning and code generation. Additionally, \citet{ruan2024observationalscalinglawspredictability} observed the log-linear relationship between the computational budget and ability factors. However, these discussions are limited to English monolingual settings, leaving cross-language generalization and scaling laws in multilingual contexts, including Japanese and English as in our study, unexplored.

\section{Experimental Settings}
\subsection{Models} \label{sec:models}
To obtain generalizable insights, we evaluated publicly available 35 Japanese, English, and Multilingual LLMs (see Table~\ref{tab:models1} in Appendix \ref{app:exp_details-models} for the complete list), which represent diverse design choices, including training data, the number of model parameters, and pre-training approach. 
The evaluated models include: English LLMs (e.g., Llama 3~\citep{metallama3}, Mistral~\citep{jiang2023mistral7b}, and Mixtral~\citep{jiang2024mixtralexperts}); 
Japanese LLMs continually pre-trained from English base LLMs on 18--175 billion tokens of Japanese text (e.g.,
Llama 3 Swallow~\citep{fujii2024continual}
and Llama 3 Youko~\citep{sawada2024release}); Japanese LLMs pre-trained primarily on 130--1,050 billion tokens of Japanese text from scratch (e.g., LLM-jp~\citep{llmjp2024llmjpcrossorganizationalprojectresearch} and Sarashina2; and multilingual LLMs pre-trained on multilingual data including Japanese (e.g., C4AI Command-R\footnote{\url{https://huggingface.co/CohereForAI/c4ai-command-r-v01}} and Qwen2~\citep{qwen2-technical-report}).
Notably, all the English LLM families that served as base models for the continually pre-trained Japanese LLMs were evaluated as well.
We focused on base models and did not evaluate instruction-tuned models to examine the effect of pre-training and avoid the confounding effects of task-oriented instruction tuning.

To estimate the computational budget for each model, we collected data on the number of model parameters and the number of training tokens for Japanese, English, and total across all languages from official sources such as technical reports, press-release documents, and model cards. Refer to Appendix \ref{app:exp_details-estimate_tokens} for details.
For a continually pre-trained model, we calculated the total number of training tokens by summing the tokens used in both initial and continual pre-training stages.

\begin{figure*}[t]
    \includegraphics[width=\linewidth]{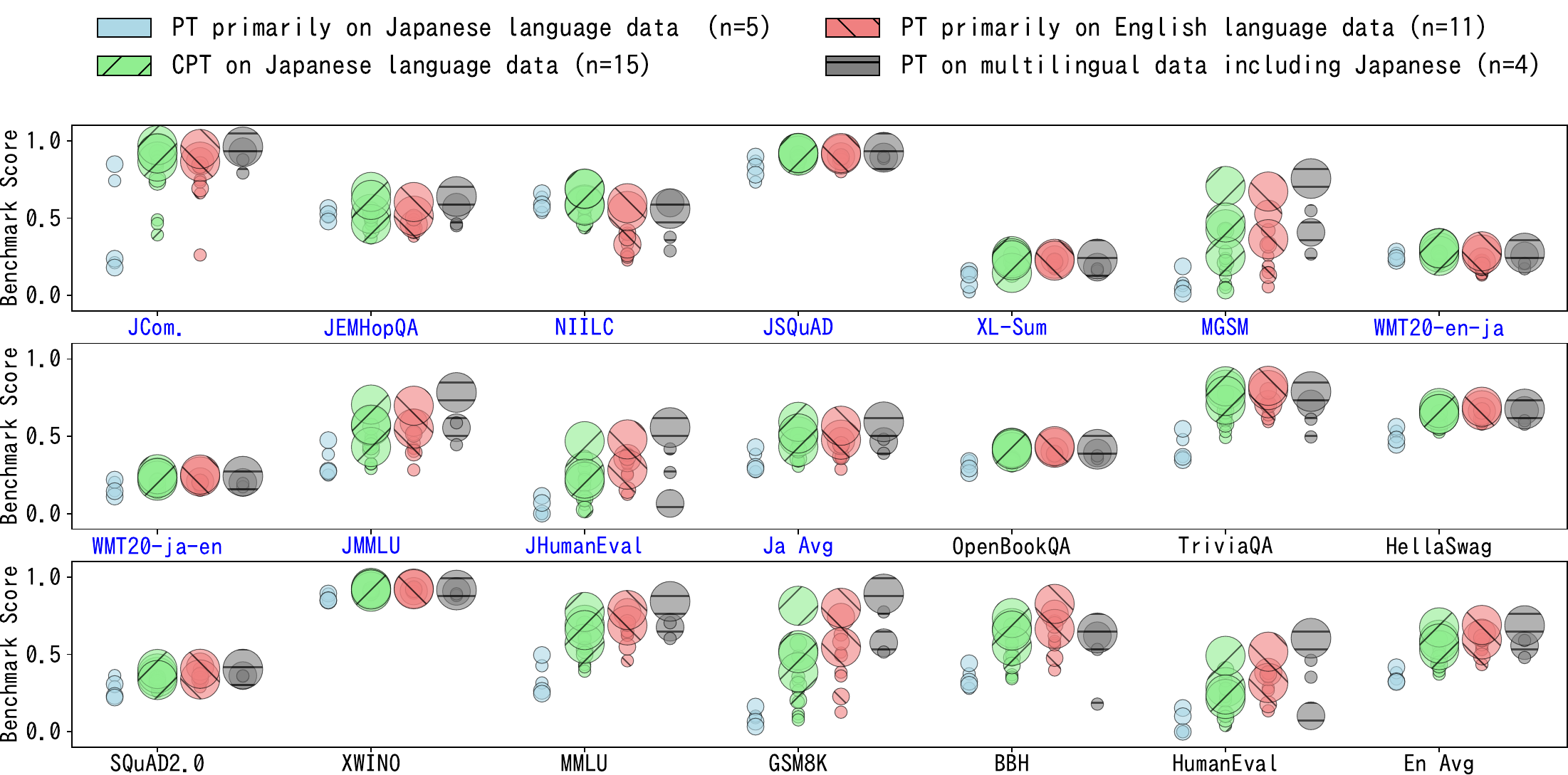}
    \caption{Task performance grouped by primary language of LLMs. Bubble size indicates the number of parameters.}
    \label{fig:eval_res}
\end{figure*}

\subsection{Evaluation Tasks and Benchmarks}
\label{sec:evaluation_tasks_and_benchmarks}
We evaluated the models using 19 evaluation benchmarks in both Japanese and English\footnote{The evaluation scores for each model
are publicly available on Zenodo, licensed under CC BY 4.0, at \url{https://doi.org/10.5281/zenodo.13160661}.
}
, which is listed in Table~\ref{tab:tasks1} of Appendix \ref{app:exp_details-tasks}.
These tasks were selected from the perspective of cross-lingual benchmarking and comprehensiveness for general-purpose LLMs.
The evaluation was conducted using zero-shot or few-shot in-context learning settings depending on tasks. Refer to Appendix \ref{app:exp_details-tasks} for details.

We employed some Japanese benchmarks corresponding to their English counterparts for cross-lingual benchmarking: code generation (JHumanEval~\citep{sato2024} vs. HumanEval~\citep{chen2021evaluatinglargelanguagemodels}), commonsense (JCommonsenseQA~\citep{kurihara-etal-2022-jglue} vs. XWINO~\citep{tikhonov-ryabinin-2021-heads} and HellaSwag~\citep{zellers-etal-2019-hellaswag}), arithmetic reasoning (MGSM~\citep{shi2023language} vs. GSM8K~\citep{cobbe2021trainingverifierssolvemath}), encyclopedic knowledge-based QA (JEMHopQA~\citep{ishii-etal-2024-jemhopqa} and NIILC~\citep{1570854175154191872}
vs. TriviaQA~\citep{joshi-etal-2017-triviaqa}),
reading comprehension (JSQuAD~\citep{kurihara-etal-2022-jglue} vs. SQuAD2~\citep{rajpurkar-etal-2018-know}), and academic subjects (JMMLU~\citep{yun2024} vs. MMLU~\citep{hendrycks2021measuring}). 
Notably, MGSM, JMMLU, and JHumanEval are translations of GSM8K, MMLU, and HumanEval, respectively. 
Cross-lingual correlations between these benchmarks provide insights into the multilinguality and language specificity of each task.
It is also worth noting that JEMHopQA and NIILC are developed based on Japanese Wikipedia and include instances that assess knowledge specific to Japanese culture, such as history, geography, notable figures and society, making them suitable for evaluating how much LLMs acquire knowledge about Japan.

For comprehensiveness, inspired by the natural language processing taxonomy~\citep{10.1145/3641289, guo2023evaluatinglargelanguagemodels} and to capture as many ability factors as possible, we included additional task benchmarks beyond cross-lingual benchmarks.
Specifically, we employed Japanese automatic summarization (XL-Sum~\citep{hasan-etal-2021-xl}), machine translation between English and Japanese (WMT20-en-ja and ja-en~\citep{barrault-etal-2020-findings}), English question answering (OpenBookQA~\citep{mihaylov-etal-2018-suit}), and logical reasoning (Big-Bench-Hard~\citep{suzgun-etal-2023-challenging}). 
Because we posit that local LLMs serve as foundational models for the target language, our evaluation focused on fundamental knowledge and skills rather than domain-specific tasks (e.g., question answering in financial or medical domains). 
Furthermore, we excluded safety and bias-related tasks, as these should be addressed in the post-training stage.

\subsection{Definition of the Computational Budgets} \label{subsec:computational_budget_definition}
The Chinchilla scaling laws~\citep{chinchilla-scaling-laws} propose an approximation for training FLOPs as $C \approx 6ND$, where $C$ represents the training FLOPs, $N$ is the number of parameters, and $D$ is the number of training tokens. Following this formula, we define $ND_l$ as the computational budget, where $D_l$ is the training tokens for the language $l$.

\subsection{Evaluation Framework and Environment} \label{sec:eval_framework_and_environment}
We evaluated all 35 LLMs on 19 task benchmarks by using a custom implementation
\footnote{Our implementation, ``swallow-evaluation'' is publicly available on our GitHub: \url{https://github.com/swallow-llm/swallow-evaluation}
\label{github_swallow-evaluation}} of existing evaluation frameworks such as llm-jp-eval~\citep{llm-jp-eval} and the Language Model Evaluation Harness\footnote{\url{https://zenodo.org/records/10256836}}. 
Refer to Table~\ref{tab:tools} for the details of implementations used for evaluation.
We used NVIDIA A100 GPUs mostly for the evaluations. 

\begin{figure*}[t]
\centering
\begin{minipage}[b]{0.585\columnwidth}
    \centering
    \includegraphics[width=0.8\linewidth]{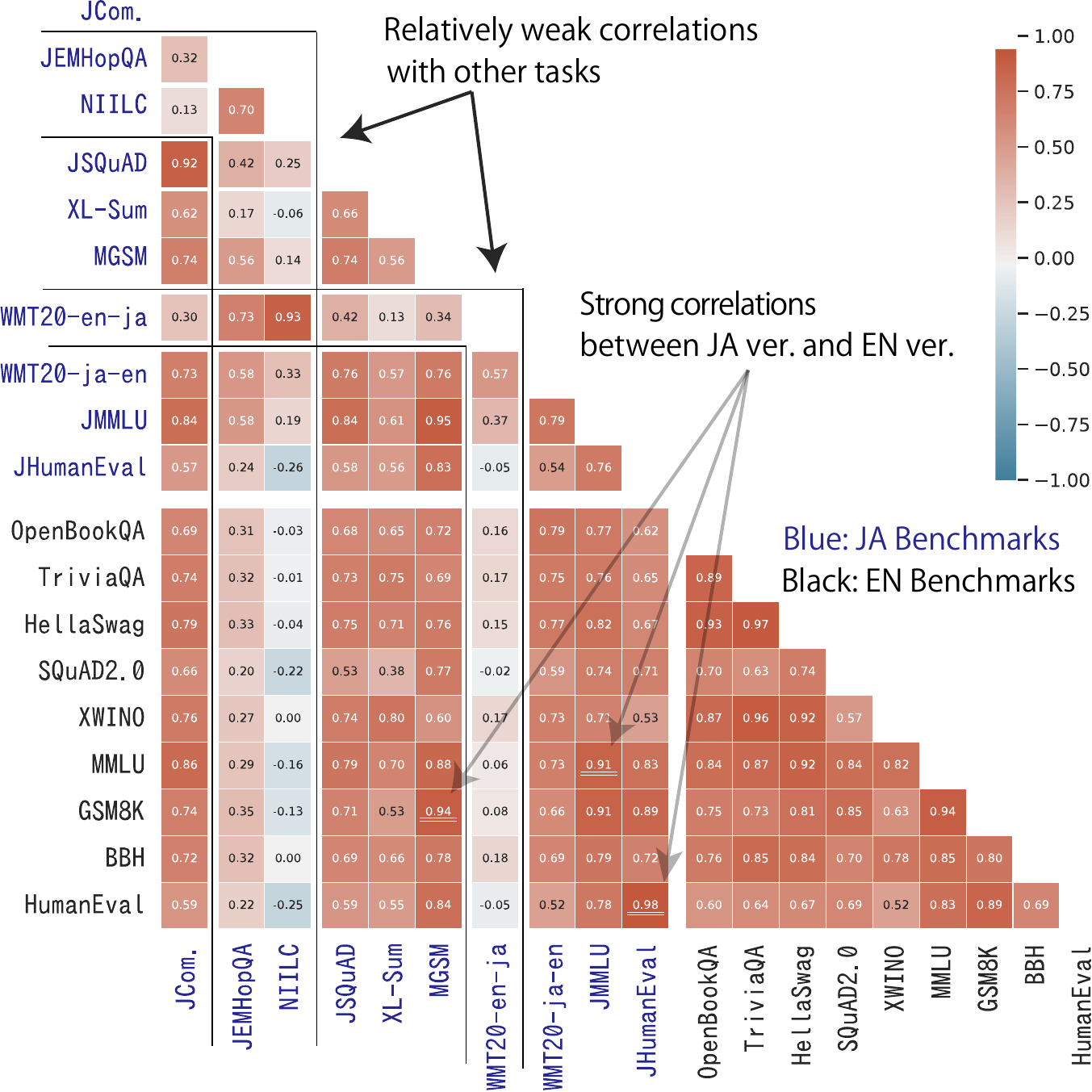}
    \caption{Pearson correlation matrix among task benchmarks ($n=35$).}
    \label{tasks_corr}
    \label{fig:tasks_corr}
\end{minipage}
\hfill
\begin{minipage}[b]{0.34\columnwidth}
    \centering
    \includegraphics[width=\columnwidth]{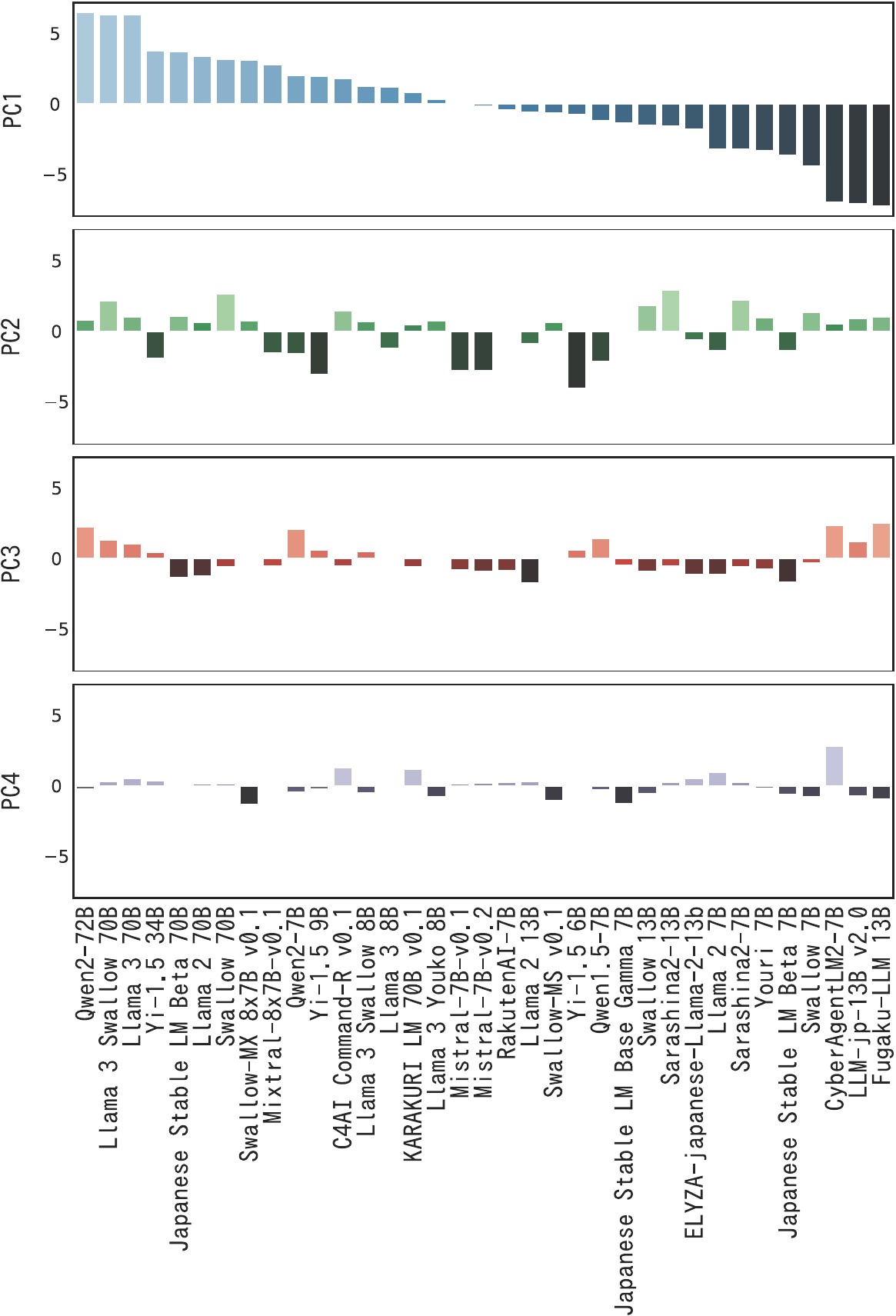}
    \caption{Principal component scores for each LLM.}
    \label{fig:pca_scores}
\end{minipage}
\end{figure*}

\section{Experimental Results} \label{sec:exp_res}
Based on the experimental setting explained in the previous section, we obtained a matrix of benchmark scores $X \in \mathbb{R}^{M \times D}$, where $M$ and $D$ are the numbers of LLMs and benchmarks, respectively ($M = 35$ and $D = 19$ in this study) and an element $X_{i,j}$ presents the score of the LLM $i$ on the benchmark $j$.
In this section, we use the benchmark scores matrix \(X\) to analyze: 1) the effects of LLM’s primary language on overall performance (\S~\ref{sec:eval_res}), 2) the similarity of benchmarks based on LLM performance (\S~\ref{sec:transferability}), 3) the ability factors of LLMs (\S~\ref{sec:PCA}), 4) whether these ability factors align with scaling laws (\S~\ref{sec:PCA_budget}), and 5) their generalizability to LLMs trained from scratch (\S~\ref{sec:pca-full-scratch}).

\subsection{Comparison of Benchmark Scores by Pre-trained Languages} \label{sec:eval_res}

Figure~\ref{fig:eval_res} presents a bubble chart showing the benchmark score distributions grouped by the primary language of the LLMs: Japanese continually pre-trained (green), Japanese trained from scratch (light blue), English (red), and Multilingual (gray).
The variable $n$ in each group represents the number of models included.

On overall, it is evident that LLMs with larger parameters tend to achieve higher scores in each group. 
When comparing benchmark scores for smaller models, there is a clear tendency for LLMs continually pre-trained on Japanese text (the green bubbles) to outperform English LLMs (the red bubbles) on Japanese benchmarks (shown in blue) except JHumanEval and MGSM. This indicates the effectiveness of continual pre-training on Japanese text.
The advantage is particularly evident in tasks such as Japanese QA (NIILC) and English-Japanese translation (WMT20-en-ja). Refer to Appendix~\ref{app:aln_eval_res} for detailed discussion.
Similarly, Japanese LLMs trained from scratch (the light blue bubbles), despite having relatively few parameters, achieve competitive scores on most Japanese benchmarks, with the exceptions of the arithmetic reasoning (MGSM) and the code-generation (JHumanEval).

\begin{figure*}[t]
    \includegraphics[width=\linewidth]{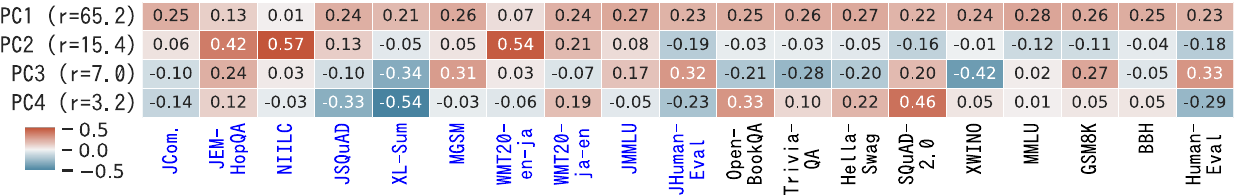}
    \caption{Factor Loadings of principal components for each benchmark ($n=35$; $r$ is the variance explained; blue: Japanese benchmarks; black: English benchmarks).}
    \label{fig:pca_factor}
\end{figure*}

\begin{figure*}[t]
    \includegraphics[width=\linewidth]{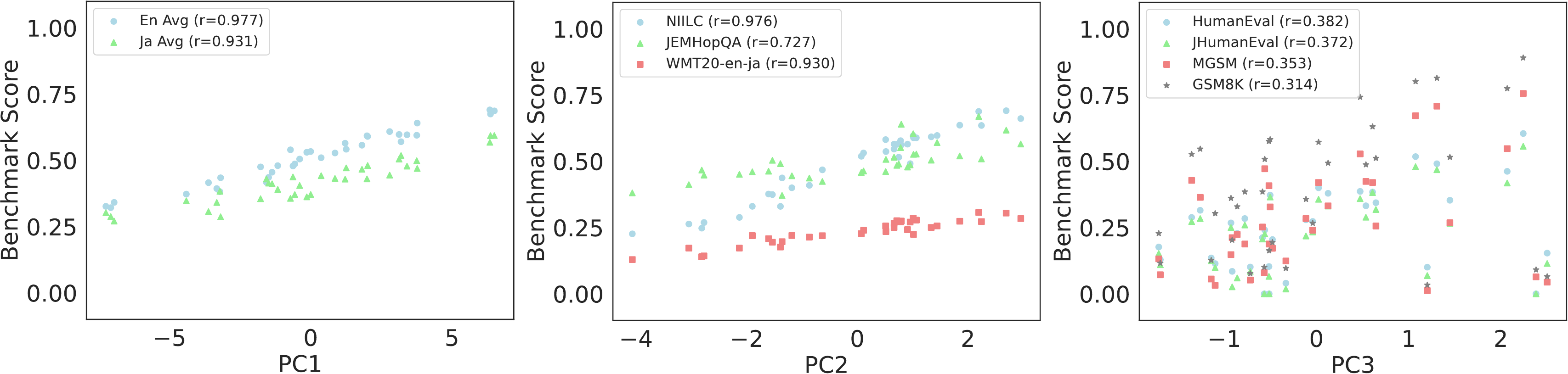}
    \caption{Relationship between principal component scores and raw benchmark scores with significant factor loadings: PC1 vs En/Ja average [left], PC2 vs Japanese knowledge-based QA and En-Ja translation [center], and PC3 vs code-generation and arithmetic reasoning [right] ($n=35$; $r$ is the pearson correlation coefficient).}
    \label{fig:pcscores_vs_benchmarkscores}
\end{figure*}

\subsection{Correlation Between Evaluation Benchmarks and Language-Specific Performance} \label{sec:transferability}
To group benchmarks based on the similarities of LLM performance, we computed a Pearson correlation between two benchmarks $a$ and $b$.
More specifically, let the column vectors $X_{:,a}$ and $X_{:,b}$ represent the array of two benchmarks $a$ and $b$, we compute the Pearson correlation $\mathrm{corr}(X_{:,a}, X_{:,b})$.
Figure~\ref{fig:tasks_corr} shows the Pearson correlation matrix, revealing two key findings\footnote{We confirmed that using Spearman's rank correlation produced no significant differences in the findings.}:

First, we observed strong cross-lingual correlations on certain tasks: academic subjects (MMLU vs. JMMLU: 0.91), arithmetic reasoning (GSM8K vs. MGSM: 0.94), and code generation (HumanEval vs. JHumanEval: 0.98). In other words, for these tasks, when LLMs perform well on the English benchmarks, they are also likely to perform well on the corresponding Japanese benchmarks. This suggests that multilinguality outweighs language specificity in these tasks, and that LLMs may generalize abilities acquired through training primarily on English text.

Second, QA tasks about Japanese knowledge (JEMHopQA, NIILC) and an English-Japanese translation task (WMT20-en-ja) exhibit relatively weak correlations with other tasks respectively.
In particular, NIILC shows negative correlations with most English tasks, and WMT20-en-ja shows almost no correlations with them.
These facts suggest that performance on these tasks may be determined by factors different from those influencing other tasks.

While we observe strong linear correlations between JMMLU, MGSM, and JHumanEval and their English counterparts, given that these are derived from English benchmarks, readers may be concerned that cross-lingual correlations of these benchmarks are overestimated. A straightforward workaround would be to evaluate using random, non-overlapping subsets of instances for each language. Instead of implementing this directly, we approximated the accuracy variation from random splits using the estimated standard error (SE) following~\citet{DBLP:journals/corr/abs-2405-14782-lm-eval-harness} and confirmed that impact of fluctuation by the SE is negligible on the observed linear trends. 
For example, MGSM has 250 instances, and the SE for an accuracy of 0.5 is approximately $\sqrt{0.5(1-0.5)/250} \approx 0.032$. In contrast, the observed standard deviation of accuracy across LLMs was 0.246, sufficiently larger than the SE.

\begin{figure*}[t]
\centering
\begin{minipage}[b]{0.475\columnwidth}
    \includegraphics[width=\linewidth]{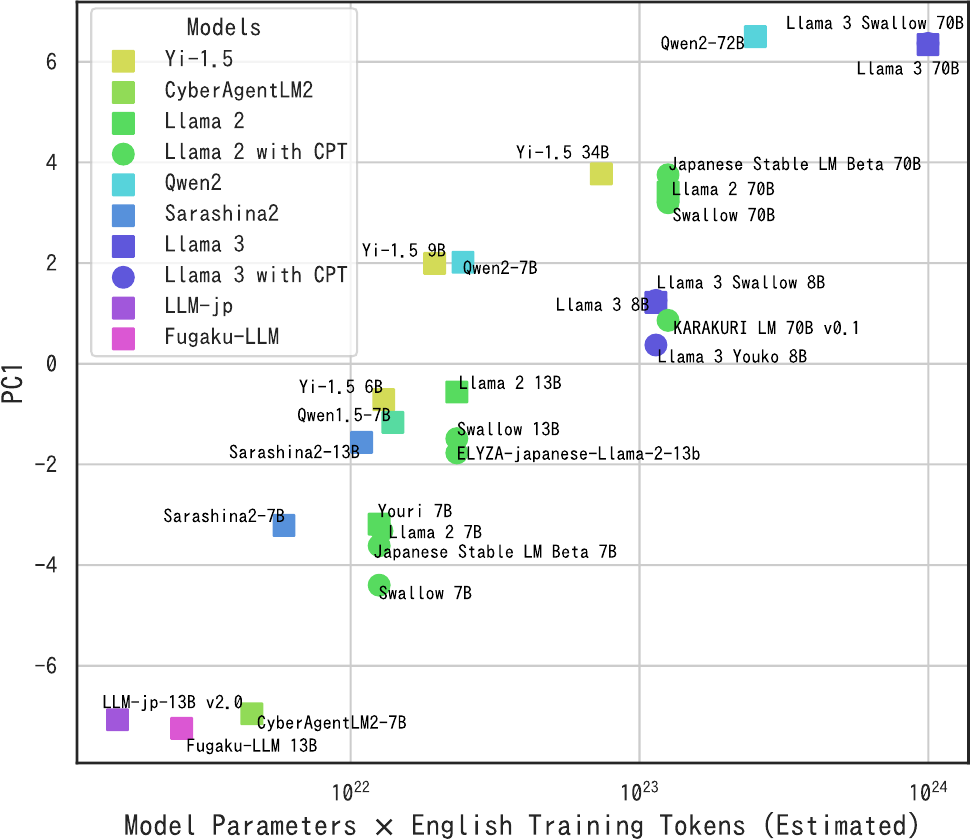}
    \caption{Relationship between the computational budget for English and PC1 scores ($n=27$).}
    \label{fig:budgetEN_pc1}
\end{minipage}
\hfill
\begin{minipage}[b]{0.475\columnwidth}
    \includegraphics[width=\linewidth]{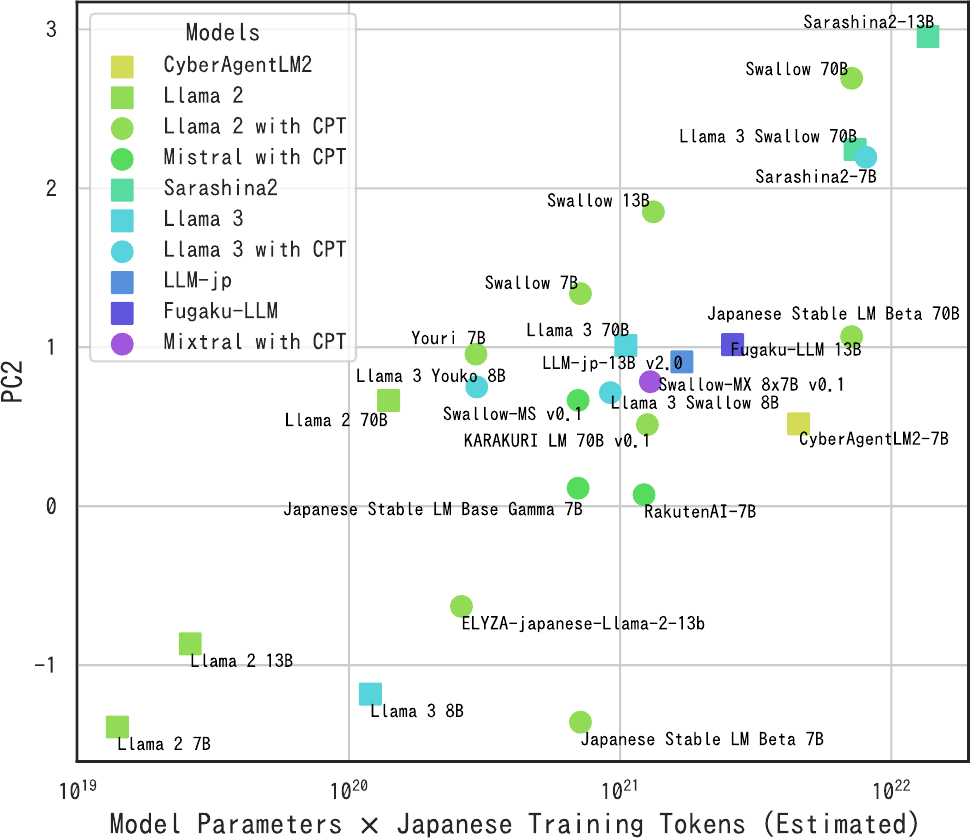}
    \caption{Relationship between the computational budget for Japanese and PC2 scores ($n=25$).}
    \label{fig:budgetJA_pc2}
\end{minipage}
\end{figure*}

\subsection{Principal Component Analysis (PCA)} \label{sec:PCA}
We observed benchmark groups from the correlation matrix in the previous subsection.
In order to identify ability factors of LLMs, we apply Principal Component Analysis (PCA)\footnote{We used the \texttt{sklearn.decomposition.PCA()} method from the \texttt{scikit-learn} package.} to project the task performance into a low-dimensional ability space.

Formally, we first standardize each column of $X$ to have mean of zero and a standard deviation of one: $\hat X$. Next, we perform eigendecomposition of the correlation matrix as $\hat X^\top \hat X = U \Lambda U^\top$, where $U = [u_1, u_2, \dots, u_D]$, and $u_j \in \mathbb{R}^{D}$ is the $j$-th unit-length eigenvector.
We then select the top four principal components (PCs), as their cumulative variance explained ($r$; contribution ratio) is 
90.8\% (= 65.2\% + 15.4\% + 7.0\% + 3.2\% from PC1 to PC4). 
We define the eigenvectors corresponding to PC1 to PC4, $U_4 = [u_1, u_2, u_3, u_4] \in \mathbb{R}^{D \times 4}$ as the factor loadings and compute corresponding PC scores as $S_4 = \hat X U_4$. Given that $U$ is an orthonormal matrix and the total variance explained by PC1--PC4 is about 90\%, the original matrix can be approximated as the product of PC scores and factor loadings: $\hat X \approx S_4 U_4^\top$.

In this way, we decompose standardized benchmark scores $\hat X$ into the product of LLM-specific principal component scores (ability factors) $S_4 \in \mathbb{R}^{M \times 4}$\ in Figure~\ref{fig:pca_scores} and benchmark-specific factor loadings $U_4 \in \mathbb{R}^{D \times 4}$ in Figure~\ref{fig:pca_factor}, which represent the associations between the ability factors and task performances\footnote{Since the signs and magnitudes of the PC scores and factor loadings are arbitrary, we adjusted the signs for ease of interpretation and normalized the factor loading vectors to have an $L_2$ norm of 1.}.

The first principal component (PC1) has relatively uniform factor loadings. As shown in Figure~\ref{fig:pcscores_vs_benchmarkscores} left, LLMs with higher PC1 scores tend to have higher average benchmark scores in both English and Japanese, suggesting that PC1 represents a general ability factor. It represents the average performance across most benchmark scores, including commonsense and reading comprehension in Japanese. This indicates that, unlike prior studies (\S~\ref{subsec:related_work_training_on_non_english_text}),  training on English text is effective and that Japanese-specific training is not necessarily for improving these abilities.

The second principal component (PC2) shows concentrated factor loadings on JEMHopQA, NIILC, and WMT20-en-ja, and relatively small factor loadings on JCommonsenseQA and JSQuAD, indicating the abilities of (encyclopedic) knowledge about Japan and English-Japanese translation. In fact, Figure~\ref{fig:pca_scores} shows that LLMs pre-trained on Japanese text, such as Swallow and Sarashina2 families, have high PC2 scores, which will be analyzed in detail in \S~\ref{sec:PCA_budget}. 
Additionally, as shown in Figure~\ref{fig:pcscores_vs_benchmarkscores} center, the higher PC2, the higher benchmark scores on those tasks. For instance, the magin of NIILC accuracy between LLMs with the lowest and highest PC2 scores is approximately 40 points.
Considering that PC1 has relatively low factor loadings for these benchmarks, PC2 represents Japanese-specific abilities, such as QA about Japanese knowledge and English-Japanese translation.
Given that PC2 strongly associates with Japanese knowledge-based QA tasks, this aligns with previous work~\citep{includeromanou2025}, which found that multilingual LLMs struggle with cultural questions, especially in languages not included in the pre-training data.

The third principal component (PC3) shows concentrated factor loadings on MGSM, GSM8K, JHumanEval, and HumanEval, representing abilities of multilingualism, language-agnostic arithmetic reasoning, and code generation.       
As shown in Figure~\ref{fig:pcscores_vs_benchmarkscores} right, 
there is a moderate trend suggesting that higher PC3 score are associated with higher benchmark scores on code-generation and arithmetic-reasoning. 

Finally, the fourth principal component (PC4) shows positive factor loadings for some English benchmarks. However, strong English LLMs, such as Llama-3-70B, do not show higher PC4 scores compared to Japanese LLMs like CyberAgentLM2-7B. In addition, given that the variance explained by PC4 is only 3.2\%, PC4 is likely to correspond to residuals that are difficult to interpret in a way tied to specific benchmarks or abilities. 

\begin{figure*}[t]
    \includegraphics[width=\linewidth]{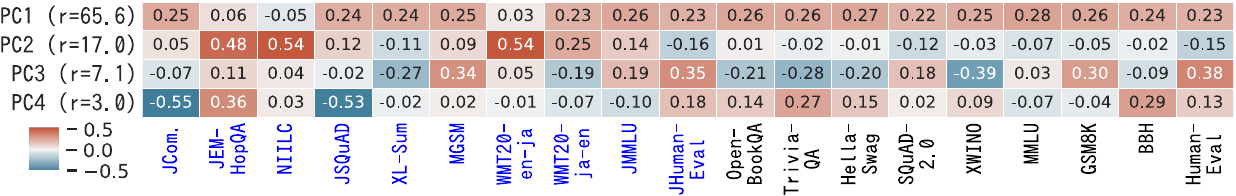}
    \caption{Factor loadings of principal components for each benchmark ($n=20$: only with models trained from scratch; $r$ is the variance explained; blue: Japanese benchmarks; black: English benchmarks).}
    \label{fig:pca_factor_wo_cpt}
\end{figure*}

\subsection{Scaling Laws between Ability Factors and Computational Budget} \label{sec:PCA_budget}
In \S~\ref{sec:PCA}, we made two key observations: 1) PC2 represents Japanese ability while PC1 represents a general ability; 2) LLMs pre-trained on Japanese text tend to have higher PC2 scores.
Based on these observations, we explore the language-specific scaling laws by examining the log-linear relationship between the computational budgets (\S~\ref{subsec:computational_budget_definition}) and principal components, which are expected to represent different abilities.

Figure~\ref{fig:budgetEN_pc1} shows the scatter plot with the English computational budget (log scale) and PC1.
It reveals that the general ability (PC1) scales with the English computational budget (Pearson's $\rho=0.916$)\footnote{The correlation with the logarithm of the total computational budget was slightly higher ($\rho=0.938$). Still, given the weak correlation with the Japanese computational budget, we concluded that it scales more with the English computational budget.}

Figure~\ref{fig:budgetJA_pc2} shows the scatter plot with the Japanese computational budget (log scale) and PC2. We can see that the Japanese ability (PC2) moderately scales with the Japanese computational budget ($\rho=0.779$). 
We also confirmed that the correlation between PC2 and the English or total computational budget is much weaker ($\rho=0.164$ and $0.186$, respectively). 
These findings indicate that PC2 and associated Japanese task performances scale with an increase in Japanese training tokens, thereby supporting our claim in \S~\ref{sec:PCA} that ``PC2 represents Japanese ability.''
Furthermore, we argue that the source of Japanese ability lies in the computational budget allocated to Japanese texts.

\subsection{PCA for LLMs Trained from Scratch} \label{sec:pca-full-scratch}
To verify that our findings are not heavily influenced by the pre-training method, we repeated the analysis after excluding continually pre-trained Japanese LLMs, retaining only 20 LLMs trained from scratch. Figure~\ref{fig:pca_factor_wo_cpt} shows the factor loadings of PCs extracted from the performance of these 20 LLMs, revealing ability factors similar to those identified in the original analysis (\S~\ref{sec:PCA}).
We omit the results of relationships between computational budgets and English and Japanese abilities, but observed the consistent correlations with Figures \ref{fig:budgetEN_pc1} and \ref{fig:budgetJA_pc2} (see Figures \ref{fig:budgetEN_f1_wo_cpt} and \ref{fig:budgetJA_f2_wo_cpt} in Appendix \ref{app:anl_build}).

\section{Conclusion and Future Work}
In this paper, we performed the most comprehensive evaluation to date, testing 35 Japanese, English, and Multilingual LLMs on 19 task benchmarks that assess the abilities in both Japanese and English.
This breadth of coverage is one of the key novelties of our study and enables us to extract more generalizable insights than prior work.
We then analyzed the cross-task and cross-lingual correlations of benchmark scores, mapped the performance in a low-dimensional ability space, and explored the relationship between ability factors and computational budgets for English and Japanese.
The correlation analysis showed strong multilingual abilities in academic subjects, code generation, and arithmetic reasoning tasks. This suggests that, in order to enhance the abilities of these tasks, there is no strong motivation for using Japanese training data.

The low-dimensional factor analysis using PCA identified three ability factors. 
PC1 represents the general ability and affects nearly all tasks except for QA about Japanese knowledge and English-Japanese translation. PC1 follows a scaling law with the computational budget for English.
Complementing PC1, PC2 represents the ability for QA about Japanese knowledge and English-Japanese translation. Interestingly, PC2 follows a scaling law with the computational budget for Japanese data.
Although PC3 represents multilingual abilities in arithmetic reasoning and code generation, we have not reached the point of identifying a scaling law that it follows.

From these analyses, we concluded that the advantage of building local LLMs by training on Japanese text is particularly evident in acquiring local knowledge written in Japanese and enhancing the ability to translate from English.
This conclusion is likely to characterize Japanese LLMs.
Our study is the first broad, unified evaluation across dozens of LLMs and an extensive benchmark suite to reveal which tasks do and do not benefit from target-language training.

We consider two directions as future work. First, we plan to extend the analysis with more LLMs and evaluation tasks to discover additional insights. This includes using LLMs with unique designs, for example, Phi family~\citep{phi_1, phi_3}, which were trained on synthetic text. We also want to add evaluation tasks such as Japanese logical reasoning and standardized admission exams.
The second direction is to extend our analysis and findings to other languages.
We believe that the conclusion of this paper can be generalized to: the advantage of building local LLMs by training in a language is acquiring local knowledge written in the language and enhancing the ability to translate from English to the language.
This direction is nontrivial because conducting LLM experiments properly requires a deep understanding of the target languages and cultures.
We hope this paper serves as a catalyst for the development and anlaysis of non-English LLMs.

\section*{Acknowledgments}
This work was supported by the \textit{ABCI Large-scale Language Model Building Support Program} of the AI Bridging Cloud Infrastructure (ABCI), built and operated by the National Institute of Advanced Industrial Science and Technology (AIST).

\section*{Ethics Statement}
This study does not evaluate the safety aspects of LLMs, such as harmlessness or honesty~\citep{askell2021generallanguageassistantlaboratory}, which are considered to be largely shaped by pre-training data. The same applies when developing local LLMs --- they are likely to absorb social group-specific biases~\citep{yanaka2024analyzingsocialbiasesjapanese}, stereotypes, and racism. Consequently, there is a concern that we may be overlooking an inconvenient side effect: it might be unavoidable for local LLMs to reinforce social biases specific to the target language.

\section*{Reproducibility Statement}
We prioritized reproducibility in our work.
As described in \S~\ref{sec:eval_framework_and_environment}, all 35 LLMs (Table~\ref{tab:models1}), 19 benchmarks (Table~\ref{tab:tasks1}), and evaluation frameworks (Table~\ref{tab:tools}) used in our study are publicly accessible.
Additionally, evaluation scores for all LLMs, along with models’ metadata--including training data, the number of model parameters, the number of training tokens, and pre-
training approach--are available (\S~\ref{sec:evaluation_tasks_and_benchmarks}) to facilitate the reproduction of statistical analyses.
Please note that our unified evaluation framework and the results are withheld here to preserve anonymity during the blind review process.
For reference, our experiments were primarily conducted on NVIDIA A100 GPUs.

\section*{Broader Impacts}
We believe our findings will contribute to the development of non-English LLMs. Moreover, this could foster a society in which every country has access to LLMs specialized in its own language and knowledge, thereby reducing the digital divide.

\bibliography{colm2025_conference}
\bibliographystyle{colm2025_conference}

\appendix
\section{Survey of Prior Work and Comparison with Our Analysis} \label{app:comp_w_prior}
We systematically surveyed prior works on non-English LLM development in two perspectives: coverage of design choices and effects of training on target languages.

At the first glance on Table~\ref{tab:prior_works}, we can find that several task types are covered sparsely.
Only 0–3 papers address machine translation (in either direction), code generation, or summarization—indicating that these areas remain largely unexplored in the literature.

More importantly, we observed the contradictory evidence for ``language-agnostic'' tasks.
The majority of prior studies actually report gains from target-language training on commonsense reasoning (8 positive, 1 neutral, 0 negative) and academic subject benchmarks (5 positive, 2 neutral, 0 negative). These findings contrast both with our results. Furthermore, there seems no clear consensus on other tasks. For reading comprehension and mathematical reasoning benchmarks, prior work offers mixed or inconclusive evidence regarding the impact of target-language data (6 positive, 3 neutral, 0 negative; 3 positive, 2 neutral, 1 negative, respectively).
 
\begin{table}[ht]
\caption{The impact of training on the target language text. $\nearrow$: Positive, $\searrow$: Negative, $\rightarrow$: Neutral, $-$: Not investigated}
\setlength{\tabcolsep}{2.5pt}
\scriptsize
\centering
\label{tab:prior_works}
\begin{tabular}{lllccccccccc}
\hline \hline
Reference & Lang & Method & \makecell[l]{Read-\\Compr.} & \makecell[l]{Com-\\mon-\\Sense\\Reason.} & \makecell[l]{Math-\\Reason.} & \makecell[l]{MT to\\Tgt Lang} & \makecell[l]{MT from\\Tgt Lang} & \makecell[l]{Acad.\\Subject} & Coding & \makecell[l]{Local\\Knowl.\\QA} & \makecell[l]{Sum-\\mar-\\iza-\\tion}\\
\hline
Ours&JA&\makecell[l]{PT\\CPT}&$\rightarrow$&$\rightarrow$&$\rightarrow$&$\nearrow$&$\nearrow$&$\rightarrow$&$\rightarrow$&$\nearrow$&$\nearrow$\\
\hline
\citet{etxaniz-etal-2024-latxa}&EU&CPT&$\nearrow$&$\nearrow$&$\nearrow,\rightarrow$&$-$&$-$&$\nearrow$&$-$&$\nearrow$&$-$\\
\citet{fujii2024continual}&JA&CPT&$\nearrow$&$\nearrow$&$\nearrow$&$\nearrow$&$\searrow$&$-$&$-$&$\nearrow$&$\searrow$\\
\citet{10799288}&TH&CPT&$\rightarrow$&$\nearrow$&$\nearrow$&$-$&$-$&$\nearrow,\rightarrow$&$-$&$-$&$-$\\
\citet{dou-etal-2024-sailor}&\makecell[l]{VI\\TH\\ID\\MS\\LO}&CPT&$\nearrow$&$\nearrow$&$-$&$-$&$-$&$\nearrow,\rightarrow$&$-$&$-$&$-$\\
\citet{joshi-etal-2025-adapting}&HI&CPT&$\nearrow$&$\nearrow$&$-$&$-$&$-$&$\nearrow$&$-$&$\nearrow$&$\nearrow$\\
\citet{vo2024redwhaleadaptedkoreanllm}&KO&CPT&$\nearrow$&$\nearrow$&$-$&$-$&$-$&$-$&$-$&$-$&$-$\\
\citet{choi-etal-2024-optimizing}&KO&CPT&$\rightarrow$&$\nearrow$&$-$&$-$&$-$&$-$&$-$&$-$&$-$\\
\citet{toraman-2024-adapting}&TR&CPT&$\rightarrow$&$\rightarrow$&$-$&$-$&$-$&$-$&$-$&$-$&$-$\\
\citet{larcher2023cabritaclosinggapforeign}&PT&CPT&$\nearrow$&$-$&$-$&$-$&$-$&$-$&$-$&$-$&$-$\\
\citet{tejaswi2024exploringdesignchoicesbuilding-arxiv}&\makecell[l]{TA\\HI\\AR\\TR}&CPT&$-$&$\nearrow$&$-$&$\nearrow$&$-$&$-$&$-$&$-$&$\nearrow$\\
\citet{cui2024efficienteffectivetextencoding}&ZH&CPT&$-$&$-$&$-$&$-$&$-$&$\nearrow$&$-$&$-$&$-$\\
\citet{NEURIPS2024_3bb42f6b}&EU&CPT&$-$&$-$&$-$&$-$&$-$&$-$&$-$&$\nearrow$&$-$\\
\citet{holmstrom-etal-2023-bridging}&SV&PT&$-$&$-$&$\searrow,\rightarrow$&$-$&$-$&$-$&$-$&$-$&$-$\\
\citet{luukkonen-etal-2023-fingpt}&FI&CPT&$-$&$-$&$-$&$-$&$-$&$-$&$-$&$-$&$-$\\
\citet{pipatanakul2023typhoonthailargelanguage}&TH&CPT&$-$&$-$&$-$&$-$&$-$&$-$&$-$&$-$&$-$\\
\hline
\end{tabular}
\end{table}

\section{Details of the Experimental Setup} \label{app:exp_details}
\subsection{Evaluated Models} \label{app:exp_details-models}
Table~\ref{tab:models1} shows a list of LLMs evaluated in this study. The table includes the name, the number of active parameters during inference, the base model from which the model was continually pre-trained, the language distribution of the training corpus, the total number of training tokens, the reported or estimated number of training tokens in English and Japanese, and the reference of each model.
\S~\ref{app:exp_details-estimate_tokens} explains the method used to estimate the number of language-specific training tokens. \textit{CPT} stands for \textit{continual pre-training}.

\setlength{\tabcolsep}{1.3pt}
\begin{table*}[t]
\caption{List of evaluated LLMs (the number of tokens is in billions [Bil], including estimates).}
\label{tab:models1}
\begin{center}
\setlength{\tabcolsep}{3pt}
\scriptsize
\begin{tabular}{l|rlllrrrp{3cm}}\hline\hline

\makecell[c]{Model name}
& \makecell[c]{Num\\of\\params} & \makecell[c]{Con-\\stru-\\ction\\met-\\hod} & \makecell[c]{Source of CPT}
& \makecell[l]{Corpus} & \makecell[c]{Training\\ tokens} 
& \makecell[c]{EN\\ tokens} & \makecell[c]{JA\\ tokens} & \makecell[c]{Reference} \\\hline

\makecell[l]{Yi-1.5 6B} & 6 & \makecell[l]{PT} & \makecell[l]{－} & \makecell[l]{ZH,EN,\\Code} & 3600 & 2170 & － & \citet{ai2024yi}	\\	
\makecell[l]{CyberAgentLM2-7B} & 7 & \makecell[l]{PT} & \makecell[l]{－} & \makecell[l]{JA,EN} & 1300 & 650 & 650 & \href{https://huggingface.co/cyberagent/calm2-7b}{cyberagent/calm2-7b}	\\	
\makecell[l]{Japanese Stable LM\\Base Gamma 7B} & 7 & \makecell[l]{CPT} & \makecell[l]{Mistral-7B-v0.1} & \makecell[l]{JA,EN} & － & － & 100 & \href{https://huggingface.co/stabilityai/japanese-stablelm-base-gamma-7b}{stabilityai/japanese-stablelm-base-gamma-7b}	\\	
\makecell[l]{Japanese StableLM\\Beta 7B} & 7 & \makecell[l]{CPT} & \makecell[l]{Llama2 7B} & \makecell[l]{JA,EN} & 2100 & 1794 & 102 & \href{https://huggingface.co/stabilityai/japanese-stablelm-base-beta-7b}{stabilityai/japanese-stablelm-base-beta-7b}	\\	
\makecell[l]{Llama 2 7B} & 7 & \makecell[l]{PT} & \makecell[l]{－} & \makecell[l]{EN} & 2000 & 1794 & 2 & \citet{metallama2}	\\	
\makecell[l]{Mistral-7B-v0.1} & 7 & \makecell[l]{PT} & \makecell[l]{－} & \makecell[l]{EN} & － & － & － & \citet{jiang2023mistral7b}	\\	
\makecell[l]{Mistral-7B-v0.2} & 7 & \makecell[l]{PT} & \makecell[l]{－} & \makecell[l]{EN} & － & － & － & \citet{jiang2023mistral7b}	\\	
\makecell[l]{Qwen1.5-7B} & 7 & \makecell[l]{PT} & \makecell[l]{－} & \makecell[l]{－} & 4000 & 2000 & － & \citet{qwen15}	\\	
\makecell[l]{Qwen2-7B} & 7 & \makecell[l]{PT} & \makecell[l]{－} & \makecell[l]{ZH,EN,\\Code+27} & 7000 & 3500 & － & \citet{qwen2-technical-report}	\\	
\makecell[l]{RakutenAI-7B} & 7 & \makecell[l]{CPT} & \makecell[l]{Mistral-7B-v0.1} & \makecell[l]{JA,EN} & － & － & 175 & \citet{rakutengroup2024rakutenai7bextendinglargelanguage}	\\	
\makecell[l]{Sarashina2-7B} & 7 & \makecell[l]{PT} & \makecell[l]{－} & \makecell[l]{JA,EN} & 2100 & 840 & 1050 & \href{https://huggingface.co/sbintuitions/sarashina2-7b}{sbintuitions/sarashina2-7b}	\\	
\makecell[l]{Swallow 7B} & 7 & \makecell[l]{CPT} & \makecell[l]{Llama2 7B} & \makecell[l]{JA,EN} & 2100 & 1794 & 102 & \citet{fujii2024continual}	\\	
\makecell[l]{Swallow-MS v0.1} & 7 & \makecell[l]{CPT} & \makecell[l]{Mistral-7B-v0.1} & \makecell[l]{JA,EN,\\Code} & － & － & 100 & \citet{fujii2024continual}	\\	
\makecell[l]{Youri 7B} & 7 & \makecell[l]{CPT} & \makecell[l]{Llama2 7B} & \makecell[l]{JA,EN} & 2040 & 1834 & 42 & \citet{sawada2024release}	\\	
\makecell[l]{Llama 3 8B} & 8 & \makecell[l]{PT} & \makecell[l]{－} & \makecell[l]{EN} & 15000 & 14250 & 15 & \citet{metallama3}	\\	
\makecell[l]{Llama 3 Swallow 8B} & 8 & \makecell[l]{CPT} & \makecell[l]{Llama3 8B} & \makecell[l]{JA,EN,\\Code} & 15100 & 14250 & 115 & \citet{fujii2024continual}	\\	
\makecell[l]{Llama 3 Youko 8B} & 8 & \makecell[l]{CPT} & \makecell[l]{Llama3 8B} & \makecell[l]{JA,EN} & 15022 & 14250 & 37 & \citet{sawada2024release}	\\	
\makecell[l]{Yi-1.5 9B} & 9 & \makecell[l]{PT} & \makecell[l]{－} & \makecell[l]{ZH,EN,\\Code} & 3100 & 2170 & － & \citet{ai2024yi}	\\	
\makecell[l]{ELYZA-japanese-\\Llama-2-13b} & 13 & \makecell[l]{CPT} & \makecell[l]{Llama2 13B} & \makecell[l]{JA} & 2018 & 1794 & 20 & \citet{elyzallama2023}	\\	
\makecell[l]{Fugaku-LLM 13B} & 13 & \makecell[l]{PT} & \makecell[l]{－} & \makecell[l]{JA,EN} & 400 & 200 & 200 & \href{https://huggingface.co/Fugaku-LLM/Fugaku-LLM-13B}{Fugaku-LLM/Fugaku-LLM-13B}	\\	
\makecell[l]{Llama 2 13B} & 13 & \makecell[l]{PT} & \makecell[l]{－} & \makecell[l]{EN} & 2000 & 1794 & 2 & \citet{metallama2}	\\	
\makecell[l]{LLM-jp-13B v2.0} & 13 & \makecell[l]{PT} & \makecell[l]{－} & \makecell[l]{JA,EN,\\Code} & 260 & 120 & 130 & \citet{llmjp2024llmjpcrossorganizationalprojectresearch}	\\	
\makecell[l]{Sarashina2-13B} & 13 & \makecell[l]{PT} & \makecell[l]{－} & \makecell[l]{JA,EN} & 2100 & 840 & 1050 & \href{https://huggingface.co/sbintuitions/sarashina2-13b}{sbintuitions/sarashina2-13b}	\\	
\makecell[l]{Swallow 13B} & 13 & \makecell[l]{CPT} & \makecell[l]{Llama2 13B} & \makecell[l]{JA,EN} & 2100 & 1794 & 102 & \citet{fujii2024continual}	\\	
\makecell[l]{Yi-1.5 34B} & 34 & \makecell[l]{PT} & \makecell[l]{－} & \makecell[l]{ZH,EN,\\Code} & 3100 & 2170 & － & \citet{ai2024yi}	\\	
\makecell[l]{C4AI Command-\\R v0.1} & 35 & \makecell[l]{PT} & \makecell[l]{－} & \makecell[l]{JA,EN,\\ZH+8} & － & － & － & \href{https://huggingface.co/CohereForAI/c4ai-command-r-v01}{CohereForAI/c4ai-command-r-v01}	\\	
\makecell[l]{Mixtral-8x7B-\\v0.1} & 13\footnotemark & \makecell[l]{PT} & \makecell[l]{－} & \makecell[l]{EN} & － & － & － & \citet{jiang2024mixtralexperts}	\\	
\makecell[l]{Swallow-MX 8x7B\\v0.1} & 13\footnotemark[\value{footnote}] & \makecell[l]{CPT} & \makecell[l]{Mixtral-8x7B-\\Instruct-v0.1} & \makecell[l]{JA,EN} & － & － & 100 & \citet{fujii2024continual}	\\	
\makecell[l]{Japanese Stable LM\\Beta 70B} & 70 & \makecell[l]{CPT} & \makecell[l]{Llama2 70B} & \makecell[l]{JA,EN} & 2100 & 1794 & 102 & \href{https://huggingface.co/stabilityai/japanese-stablelm-base-beta-70b/discussions}{stabilityai/japanese-stablelm-base-beta-70b}	\\	
\makecell[l]{KARAKURI LM 70B\\v0.1} & 70 & \makecell[l]{CPT} & \makecell[l]{Llama2 70B} & \makecell[l]{JA,EN} & 2016 & 1794 & 18 & \citet{karakuri_lm_70b_v01}	\\	
\makecell[l]{Llama 2 70B} & 70 & \makecell[l]{PT} & \makecell[l]{－} & \makecell[l]{EN} & 2000 & 1794 & 2 & \citet{metallama2}	\\	
\makecell[l]{Llama 3 70B} & 70 & \makecell[l]{PT} & \makecell[l]{－} & \makecell[l]{EN} & 15000 & 14250 & 15 & \citet{metallama3}	\\	
\makecell[l]{Llama 3 Swallow 70B} & 70 & \makecell[l]{CPT} & \makecell[l]{Llama3 70B} & \makecell[l]{JA,EN,\\Code} & 15100 & 14250 & 115 & \citet{fujii2024continual}	\\	
\makecell[l]{Swallow 70B} & 70 & \makecell[l]{CPT} & \makecell[l]{Llama2 70B} & \makecell[l]{JA,EN} & 2100 & 1794 & 102 & \citet{fujii2024continual}	\\	
\makecell[l]{Qwen2-72B} & 72 & \makecell[l]{PT} & \makecell[l]{－} & \makecell[l]{ZH,EN,\\Code+27} & 7000 & 3500 & － & \citet{qwen2-technical-report}	\\	\hline
\end{tabular}
\end{center}
\end{table*}

\footnotetext{Number of active parameters on inference. The total number of parameters is 47B.}
\addtocounter{footnote}{-2}

\subsection{Evaluation Tasks and Benchmarks} \label{app:exp_details-tasks}
Table \ref{tab:tasks1} provides an overview of the evaluation benchmarks used in this study. The table includes the benchmark name, a brief description, the language of the task, the metric for scoring the model's output, the experimental setting (e.g., few-shot, zero-shot, chain-of-thought), and the reference of each benchmark. The scale of evaluation metrics is normalized between 0 and 1, and \textit{EM} means \textit{exact match}.

\begin{table*}[t]
\caption{List of benchmarks used for evaluation.}
\label{tab:tasks1}
\hbox to\hsize{\hfil
\scriptsize
\begin{tabular}{l|llllp{3.5cm}}\hline\hline

	Name  & 	Description  & Lang.	 & 	\makecell[c]{Eval.\\metric\footnotemark$^{,}$\footnotemark}	 & 	\makecell[c]{Exp.\\setup}  & 	Reference \\\hline
\makecell[l]{JcommonsenseQA\\(JCom.)}	 & \makecell[l]{Multiple-choice questions\\with 5 options based on\\a knowledge base}	 & JA	 & \makecell[l]{Acc.}	 & \makecell[l]{4-shot}	 & \citet{kurihara-etal-2022-jglue}	\\
\makecell[l]{JEMHopQA}	 & \makecell[l]{Free-form question answering\\to evaluate knowledge\\and reasoning ability}	 & JA	 & \makecell[l]{Char F1}	 & \makecell[l]{4-shot}	 & \citet{ishii-etal-2024-jemhopqa}	\\
\makecell[l]{NIILC}	 & \makecell[l]{Free-form question answering\\where answers can be obtained\\from an encyclopedia}	 & JA	 & \makecell[l]{Char F1}	 & \makecell[l]{4-shot}	 & \citet{1570854175154191872}	\\
\makecell[l]{JSQuAD}	 & \makecell[l]{Free-form question answering\\on Wikipedia articles}	 & JA	 & \makecell[l]{Char F1}	 & \makecell[l]{4-shot}	 & \citet{kurihara-etal-2022-jglue}	\\
\makecell[l]{XL-Sum}	 & \makecell[l]{Generating summaries\\from BBC articles}	 & JA	 & \makecell[l]{ROUGE-2}	 & \makecell[l]{1-shot}	 & \citet{hasan-etal-2021-xl}	\\
\makecell[l]{MGSM}	 & \makecell[l]{Japanese translation of the\\primary school math\\word problem\\dataset (GSM8K)}	 & JA	 & \makecell[l]{Acc.\\(EM)}	 & \makecell[l]{4-shot}	 & \citet{shi2023language}	\\
\makecell[l]{WMT20(en-ja)}	 & \makecell[l]{English-Japanese translation\\of news articles}	 & JA	 & \makecell[l]{BLEU}	 & \makecell[l]{4-shot}	 & \citet{barrault-etal-2020-findings}	\\
\makecell[l]{WMT20(ja-en)}	 & \makecell[l]{Japanese-to-English translation\\of news articles}	 & JA	 & \makecell[l]{BLEU}	 & \makecell[l]{4-shot}	 & \citet{barrault-etal-2020-findings}	\\
\makecell[l]{JMMLU}	 & \makecell[l]{Japanese translation of the\\multiple-choice benchmark\\MMLU (53 subjects)}	 & JA	 & \makecell[l]{Acc.}	 & \makecell[l]{5-shot}	 & \citet{yun2024}	\\
\makecell[l]{JHumanEval}	&	\makecell[l]{Japanese translation of 
\\HumanEval}	&	JA	&	\makecell[l]{pass@1}	&	\makecell[l]{0-shot\\10 trials}	&	\citet{sato2024}	\\
\makecell[l]{OpenBookQA}	 & \makecell[l]{Multiple-choice questions based\\on scientific knowledge and\\common sense}	 & EN	 & \makecell[l]{Acc.}	 & \makecell[l]{4-shot}	 & \citet{mihaylov-etal-2018-suit}	\\
\makecell[l]{TriviaQA}	 & \makecell[l]{Free-form question answering\\based on trivia knowledge}	 & EN	 & \makecell[l]{Acc.\\(EM)}	 & \makecell[l]{4-shot}	 & \citet{joshi-etal-2017-triviaqa}	\\
\makecell[l]{HellaSwag}	 & \makecell[l]{Multiple-choice questions\\to predict the next event}	 & EN	 & \makecell[l]{Acc.}	 & \makecell[l]{4-shot}	 & \citet{zellers-etal-2019-hellaswag}	\\
\makecell[l]{SQuAD2}	 & \makecell[l]{Free-form question answering\\based on a supporting document}	 & EN	 & \makecell[l]{Acc.\\(EM)}	 & \makecell[l]{4-shot}	 & \citet{rajpurkar-etal-2018-know}	\\
\makecell[l]{XWINO}	 & \makecell[l]{Binary-choice questions\\to identify the antecedent\\of a pronoun in a sentence}	 & EN	 & \makecell[l]{Acc.}	 & \makecell[l]{4-shot}	 & \citet{tikhonov-ryabinin-2021-heads}	\\
\makecell[l]{MMLU}	 & \makecell[l]{Multiple-choice questions\\across 57 subjects}	 & EN	 & \makecell[l]{Acc.}	 & \makecell[l]{5-shot}	 & \citet{hendrycks2021measuring}	\\
\makecell[l]{GSM8K}	 & \makecell[l]{Primary school math word\\problem dataset}	 & EN	 & \makecell[l]{Acc.\\(EM)}	 & \makecell[l]{4-shot}	 & \citet{cobbe2021trainingverifierssolvemath}	\\
\makecell[l]{BBH}	 & \makecell[l]{23 challenging tasks from\\the BIG-Bench dataset}	 & EN	 & \makecell[l]{Acc.\\(EM)}	 & \makecell[l]{3-shot\\CoT}	 & \citet{suzgun-etal-2023-challenging}	\\
\makecell[l]{HumanEval}	 & \makecell[l]{Evaluation of code generation\\ability via unit tests}	 & EN	 & \makecell[l]{pass@1}	 & \makecell[l]{0-shot\\10 trials}	 & \citet{chen2021evaluatinglargelanguagemodels}	\\\hline

 \end{tabular}\hfil}
\end{table*}

\subsection{Estimating the Number of Training Tokens} \label{app:exp_details-estimate_tokens}
The numbers of language-specific training tokens (in billions) were either obtained from or calculated based on official sources such as technical reports, release documents, or model cards. When an exact number was unavailable in the source, we used the following estimates:

\begin{itemize}
    \item Ratio of Japanese training tokens:
    \begin{itemize}
        \item Llama 2, Llama 3: 0.1\%
        \item Mistral, Mixtral: 0\%
        \item Full-scratch Japanese LLMs: 50\%
        \item Japanese LLMs with CPT: 100\%
    \end{itemize}
\end{itemize}

\begin{itemize}
    \item Ratio of English training tokens:
    \begin{itemize}
        \item Qwen1.5, Qwen2: 50\%
        \item Yi-1.5: 70\%
        \item Llama 2: 89.7\%
        \item Llama 3: 95\%
    \end{itemize}
\end{itemize}

A symbol `–' in Table~\ref{tab:models1} indicates that the number could not be obtained or estimated despite our best efforts. We excluded these LLMs from the analysis of the scaling laws in \S~\ref{sec:PCA_budget}.

\subsection{Evaluation Framework} \label{app:exp_details-tools}
Table~\ref{tab:tools} reports a list of evaluation frameworks used in this study. 
The table shows the framework name, a brief description, and the reference of the framework.
We slightly customized these evaluation frameworks to cover benchmarks that are not officially supported and to implement workarounds for LLMs; for example, some LLMs require special tokens or line breaks in the prompt to generate valid outputs.

\begin{table*}[t]
\caption{List of evaluation frameworks.}
\label{tab:tools}
\hbox to\hsize{\hfil
\scriptsize
\begin{tabular}{l|ll}\hline\hline

	Name  & 	Description  & Reference	\\\hline

\makecell[l]{LLM-jp eval\\(1.3.0)}	 & \makecell[l]{Automatic evaluation tool\\for Japanese LLMs}	 & \citet{llm-jp-eval}	\\  \\
\makecell[l]{JP Language Model\\Evaluation Harness\\(commit \#9b42d41)}	 & \makecell[l]{An evaluation framework\\for Japanese LLMs}	 & \href{https://zenodo.org/records/10256836}{zenodo.10256836}	\\ \\
\makecell[l]{Language Model\\Evaluation Harness\\(0.4.2)}	 & \makecell[l]{An evaluation framework\\for LLMs}	 & \href{https://zenodo.org/records/10256836}{zenodo.10256836}	\\ \\
\makecell[l]{Code Generation LM\\Evaluation Harness\\(commit \#0261c52)}	 & \makecell[l]{An evaluation framework\\for code generation task}	 & \citet{bigcode-evaluation-harness}	\\ \\

\hline

\end{tabular}\hfil}
\end{table*}

\subsection{Details of LLM Grouping} \label{app:exp_details-eval_res_models}
Table~\ref{tab:eval_res_models} shows the breakdown of LLM groups used in Figure~\ref{fig:eval_res}.

\begin{table*}[t]
    \caption{Breakdown of LLM groups used in Figure~\ref{fig:eval_res}.}
    \label{tab:eval_res_models}
    \hbox to\hsize{\hfil
\scriptsize
\begin{tabular}{l|ll}\hline\hline

	Category  & 	Models		\\\hline
\makecell[l]{Japanese LLMs pre-trained\\from scratch}	 & \makecell[l]{CyberAgentLM2-7B， Sarashina2-7B， Sarashina2-13B，\\Fugaku-LLM 13B， LLM-jp-13B v2.0}	\\ \\
\makecell[l]{LLMs continually pre-trained\\on Japanese text}	 & \makecell[l]{Japanese Stable LM Base Gamma 7B\\Japanese Stable LM Beta 7B，\\RakutenAI-7B， Swallow 7B， Swallow-MS v0.1，\\Youri 7B， Llama 3 Swallow 8B，\\Llama 3 Youko 8B， ELYZA-japanese-Llama-2-13b，\\Swallow 13B， Swallow-MX 8x7B v0.1，\\Japanese Stable LM Beta 70B， KARAKURI LM 70B v0.1，\\Llama 3 Swallow 70B， Swallow 70B}	\\ \\
\makecell[l]{Egnlish LLMs}	 & \makecell[l]{Yi-1.5 6B， Llama 2 7B， Mistral-7B-v0.1，\\Mistral-7B-v0.2， Llama 3 8B， Yi-1.5 9B，\\Llama 2 13B， Yi-1.5 34B， Mixtral-8x7B-v0.1，\\Llama 2 70B， Llama 3 70B}	\\ \\
\makecell[l]{Multilingual LLMs}	 & \makecell[l]{C4AI Command-R v0.1,\\Qwen1.5-7B， Qwen2-7B， Qwen2-72B}	\\\hline

\end{tabular}\hfil}
\end{table*}

\section{Analysis of the Evaluation Results} \label{app:aln_eval_res}
This section presents detailed observations that complement the explanation in \S~\ref{sec:eval_res}.

\subsection{Performance Difference between the Pre-trained Languages}
Figure~\ref{fig:eval_res} reveals a notable observation: the scores of Japanese LLMs pre-trained from scratch (the blue box) are consistently lower than those of continually pre-trained models. This may be due to the relatively small number of parameters of the LLMs in this category (e.g. CyberAgentLM2-7B, Sarashina2-7B, Fugaku-LLM 13B), as well as the limited training budget (i.e., number of training tokens) available for developing LLMs from scratch. This highlights a challenge in developing local LLMs in Japan.

Additionally, compared to other groups, multilingual LLMs (the black box) performed significantly better in arithmetic reasoning (MGSM and GSM8K) and code generation (JHumanEval and HumanEval) tasks. However, we believe that this does not reflect the overall strength of multilingual LLMs, but rather the strengths of Qwen family~\citep{qwen2-technical-report}, which represents three out of four LLMs in this group.

\subsection{Variations in Task Scores}
Figure \ref{fig:eval_res} highlights tasks with both high and low score variances. Tasks with low score variances can be grouped into two categories:
\begin{enumerate}
    \item Benchmarks evaluated with n-gram based metrics (e.g. WMT20-ja-en and WMT20-en-ja with BLEU, and XL-Sum with ROUGE-2).
    \item Tasks requiring essential skills (e.g. JSQuAD and SQuAD2.0 (reading comprehension), and OpenBookQA and XWINO (commonsense)).
\end{enumerate}

In contrast, tasks with high score variances can be grouped into two categories:
\begin{enumerate}
    \item Tasks requiring specific capabilities (e.g. MGSM, GSM8K (arithmetic reasoning), JHumanEval and HumanEval (code generation))
    \item Knowledge-intensive tasks (e.g. NIILC, JMMLU, MMLU, and TriviaQA)
\end{enumerate}
The scores for these tasks heavily depend on whether a model possesses the necessary capabilities or specialized knowledge, which leads to a greater variance.

\section{Robustness Check of Findings Obtained from Experimental Results}
To test the robustness of the findings presented in \S~\ref{sec:exp_res}, we conducted two additional analyses using different methods and settings: the use of maximum likelihood estimation and Promax rotation\footnote{We used the \texttt{factor\_analyzer.FactorAnalyzer()} and \texttt{factor\_analyzer.Rotator()} method from the \texttt{factor\_analyzer} package.} instead of PCA (in \S~\ref{sec:PCA}); and exclusion of continually pre-trained models to focus on models trained from scratch.
Moreover, we performed leave-one-out cross-validation to confirm that our insights derived from observational approach are robust to statistical errors.

\subsection{Maximum Likelihood Estimation and Promax Rotation} \label{app:max_likelihood_estimation_promax_rotation}
\begin{figure*}[t]
    \includegraphics[width=\linewidth]{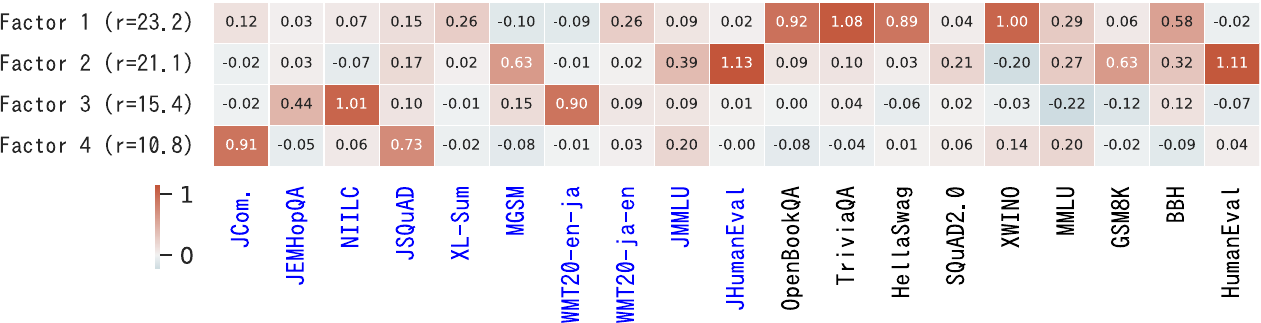}
    \caption{Factor loadings by task with Promax rotation applied ($n=35$; $r$ represents a contribution; blue and black colors correspond to Japanese and English task benchmarks, respectively).}
    \label{fig:anl_promax_loadings}
\end{figure*}

Figure~\ref{fig:anl_promax_loadings} presents factor loadings with Promax rotation applied. This figure reveals two factors similar to those identified in \S~\ref{sec:PCA}: ability factor for arithmetic reasoning and code generation (Factor 2 for PC3), and ability factor Japanese (Factor 3 for PC2).
In contrast, the first factor (Factor 1) seems to represent English ability, not the general ability (PC1), since the loading scores are strongly positive on the English task benchmarks such as OpenBookQA, TriviaQA, HellaSwag, and XWINO.

\begin{figure*}[t]
    \includegraphics[width=\linewidth]{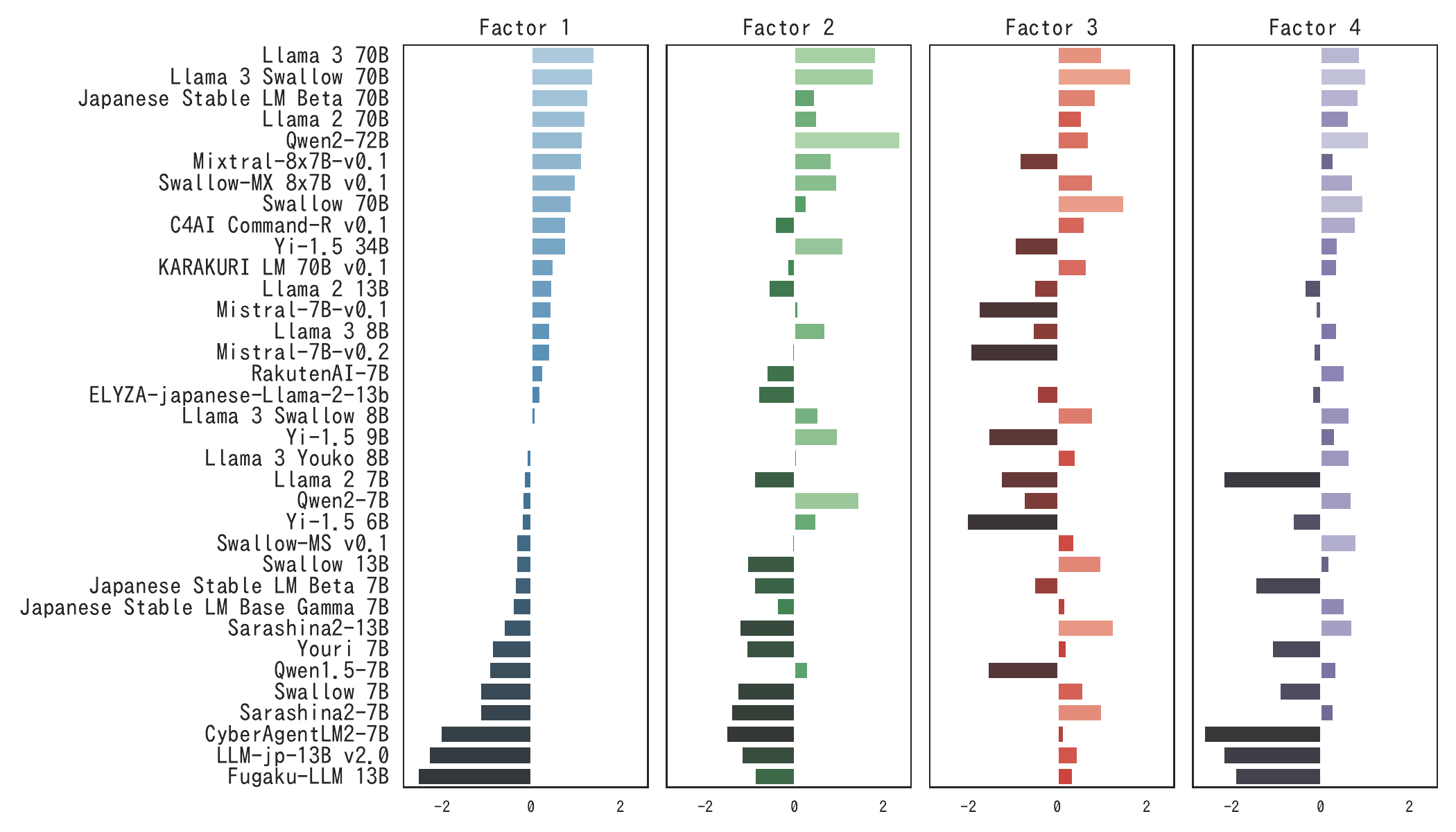}
    \caption{Factor scores for each model with Promax rotation applied.}
    \label{fig:anl_promax_pca_scores}
\end{figure*}

Additionally, the fourth factor (Factor 4) seems to be a distinct ability factor for Japanese at first glance since the loading scores are strongly positive on two Japanese task benchmarks (JCom. and JSQuAD). However, the correlation coefficient with the logarithm of the computational budget for Japanese is as small as 0.241, much lower than that of the computational budget for English (0.788). Figure~\ref{fig:anl_promax_pca_scores} shows small Factor 4 scores on Japanese LLMs, such as Llama 3 Youko 8B, Japanese Stable LM Beta 7B, CyberAgentLM2-7B, LLM-jp-13B v2.0 and Fugaku-LLM 13B. Even strong Japanese LLMs (e.g., Llama 3 Swallow 70B, Japanese Stable LM Base Gamma 7B) do not show high scores compared to non-Japanese LLMs. Therefore, the fourth factor should be considered as a residual that is difficult to interpret; therefore, commonsense tasks and reading comprehension do not determine Japanese abilities.

To sum, these results confirm two similar factors to those identified in \S~\ref{sec:PCA} (an ability factor for arithmetic reasoning and code generation, and a Japanese ability factor) and two unique factors (an English ability factor and a residual factor).

\begin{figure*}[t]
\centering
\begin{minipage}[b]{0.475\columnwidth}
    \includegraphics[width=\linewidth]{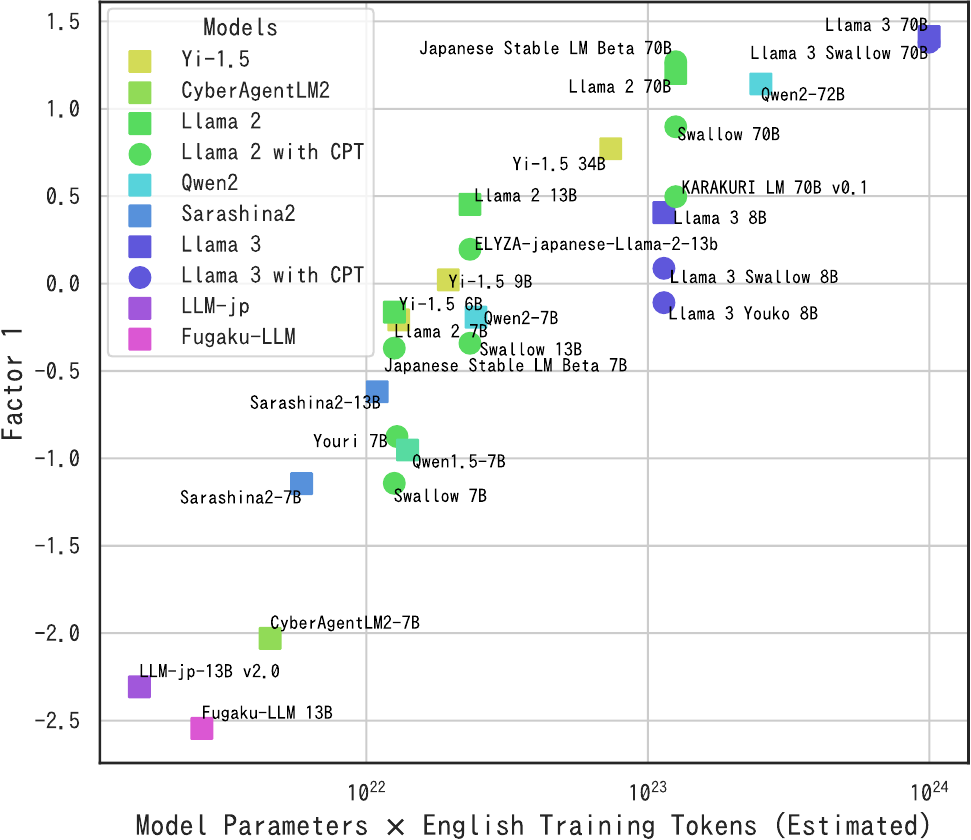}
    \caption{Relationship between the computational budget for English and Factor 1 ($n=27$).}
    \label{fig:budgetEN_f1}
\end{minipage}
\hfill
\begin{minipage}[b]{0.475\columnwidth}
    \includegraphics[width=\linewidth]{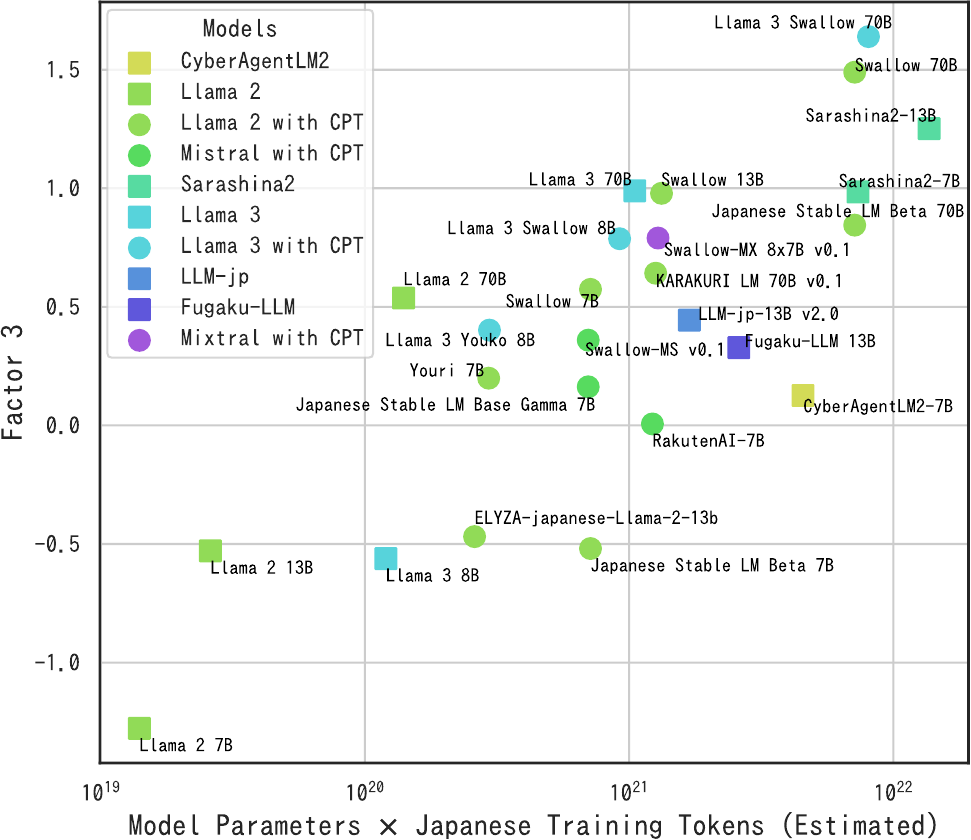}
    \caption{Relationship between the computational budget for Japanese and Factor 3 ($n=27$).}
    \label{fig:budgetJA_f3}
\end{minipage}
\\
\begin{minipage}[b]{0.475\columnwidth}
    \includegraphics[width=\linewidth]{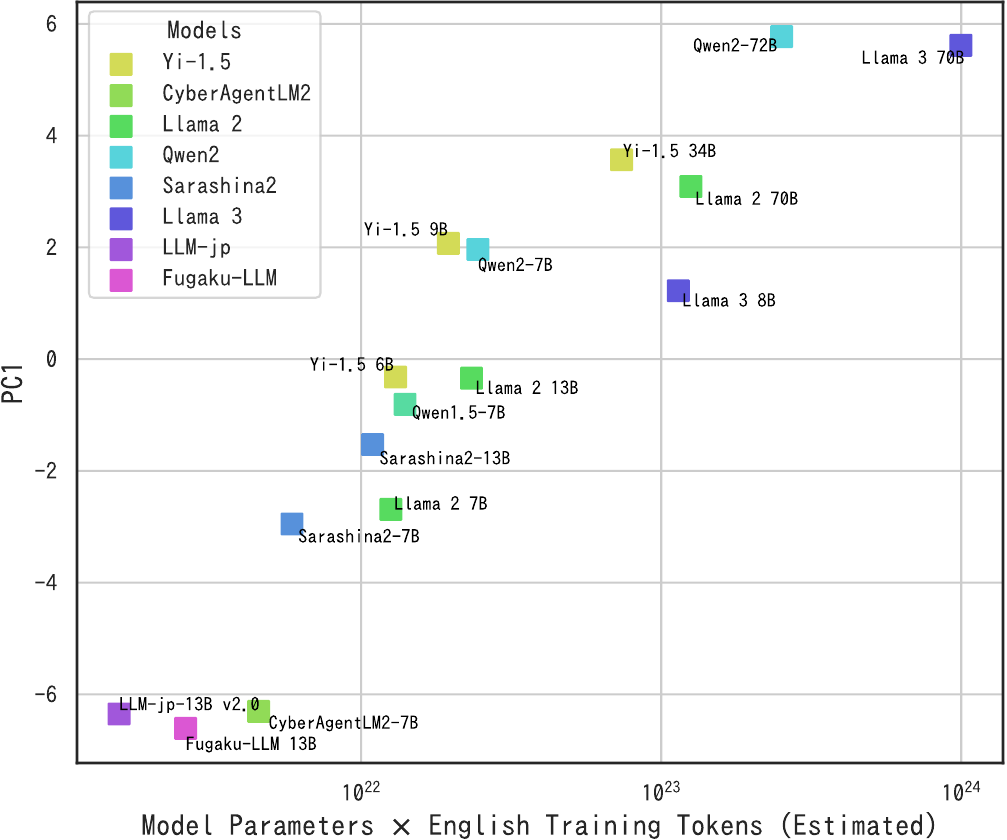}
    \caption{Relationship between the computational budget for English and PC1 ($n=16$; only with models trained from scratch).}
    \label{fig:budgetEN_f1_wo_cpt}
\end{minipage}
\hfill
\begin{minipage}[b]{0.475\columnwidth}
    \includegraphics[width=\linewidth]{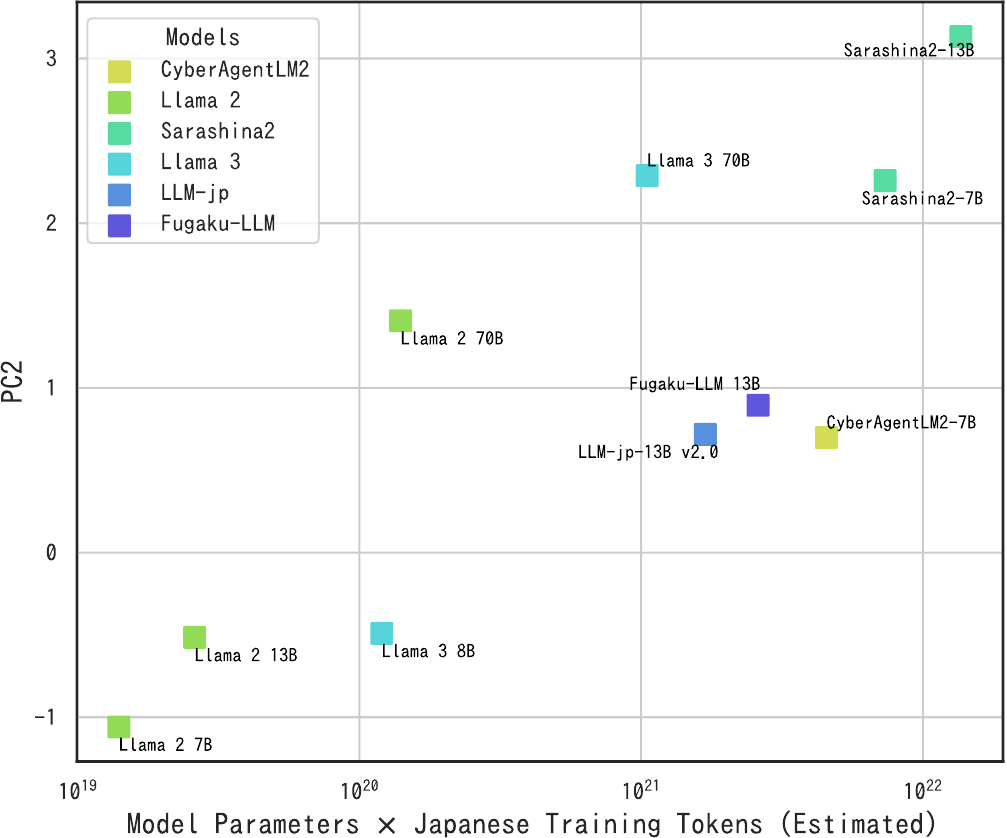}
    \caption{Relationship between the computational budget for Japanese and PC2 ($n=10$; only with models trained from scratch).}
    \label{fig:budgetJA_f2_wo_cpt}
\end{minipage}
\end{figure*}

\subsection{Analysis with only Full-scratch Models} \label{app:anl_build}
We removed continually pre-trained LLMs, which are categorized as \textit{LLMs continually pre-trained on Japanese text} in Table~\ref{tab:eval_res_models} and conducted the same analysis as in \S~\ref{sec:transferability} to \S~\ref{sec:PCA_budget}.

\begin{figure*}[t]
\centering
\begin{minipage}[b]{0.625\columnwidth}
    \centering
    \includegraphics[width=\linewidth]{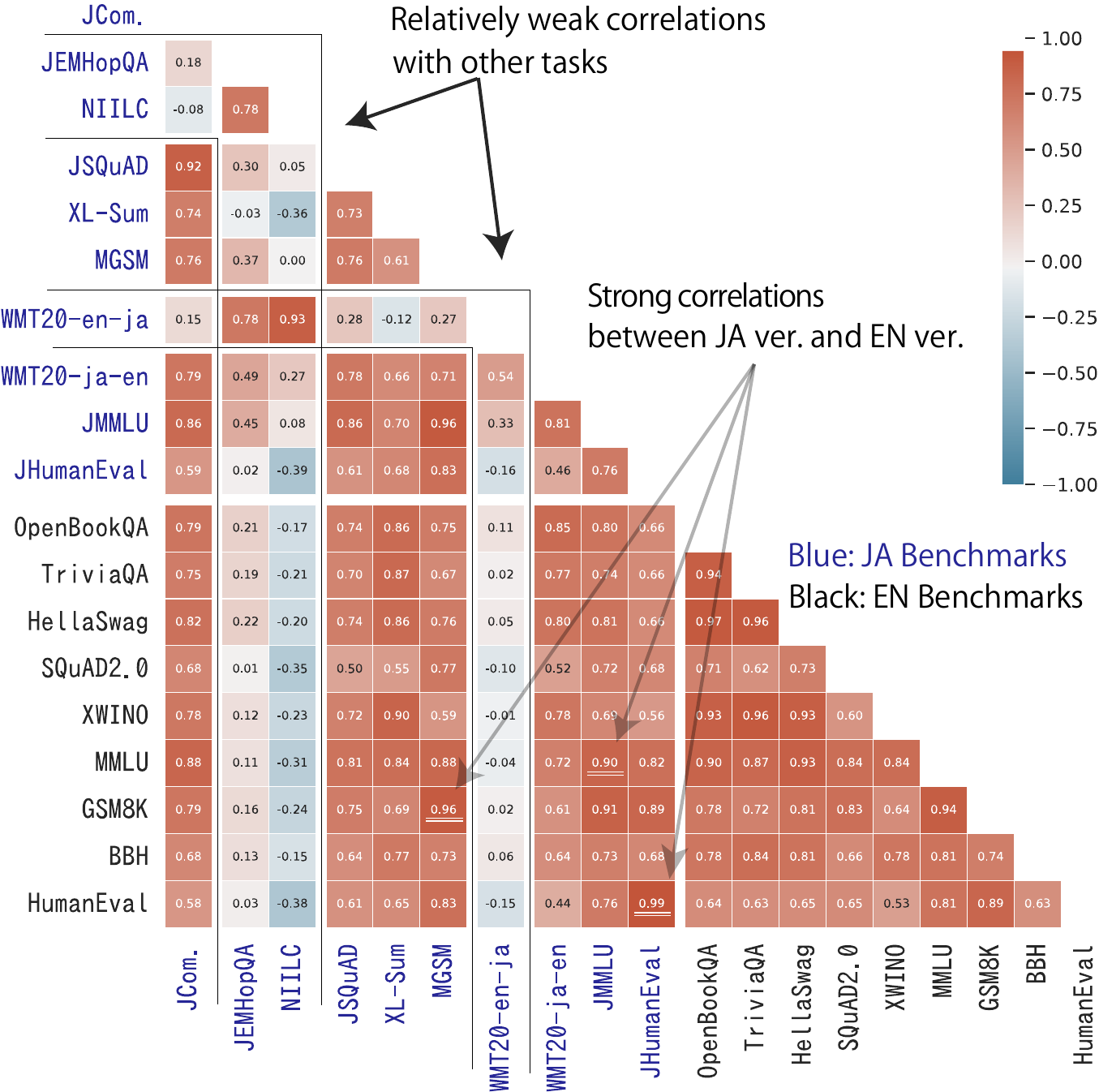}
    \caption{Pearson correlation matrix among benchmark scores ($n=20$; only with models trained from scratch).}
    \label{fig:tasks_corr_wo_cpt}
\end{minipage}
\hfill
\begin{minipage}[b]{0.325\columnwidth}
    \includegraphics[width=\linewidth]{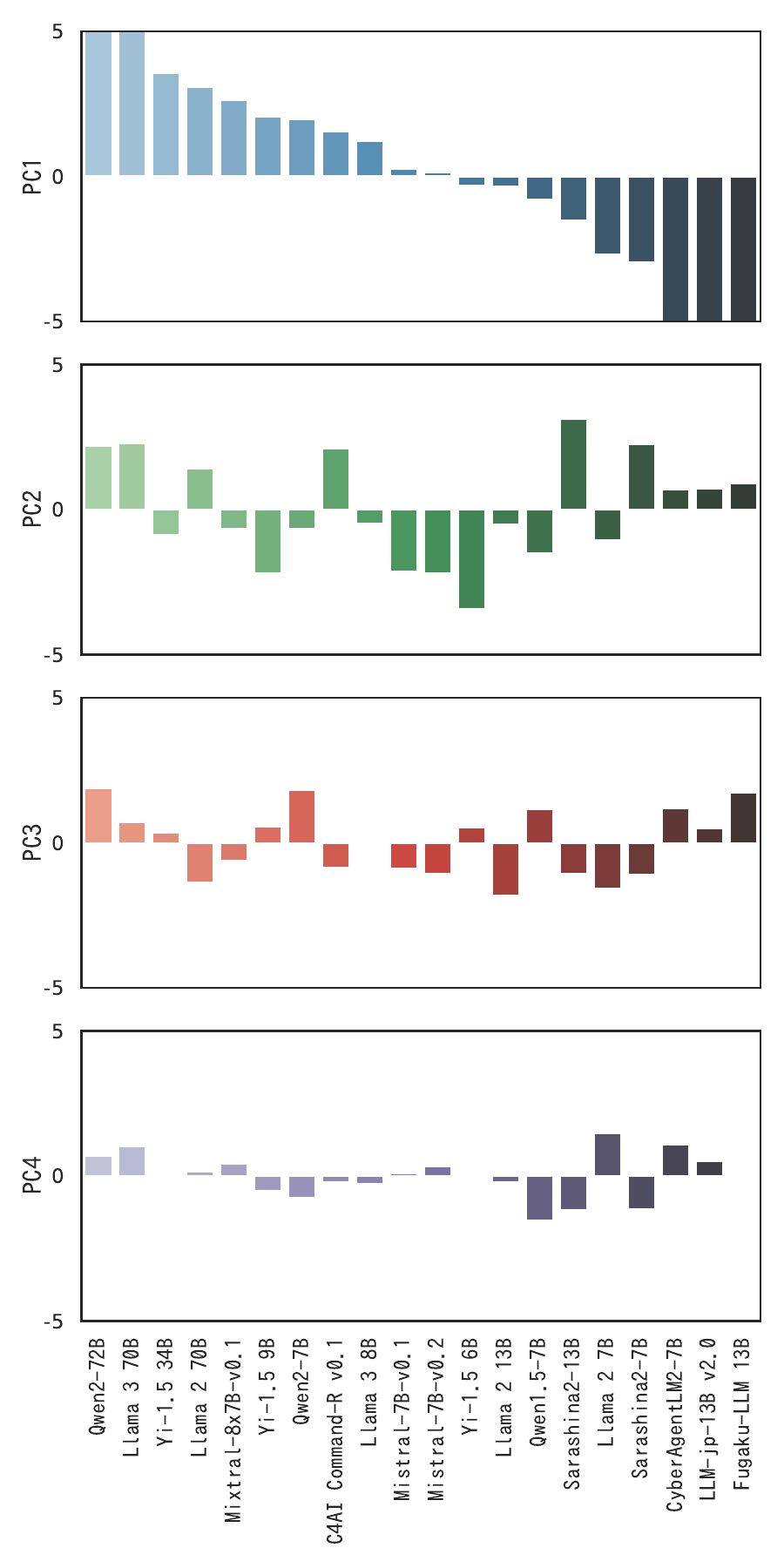}
    \caption{Principal component scores for each model ($n=20$; only with models trained from scratch).}
    \label{fig:pca_factor_score_wo_cpt}
\end{minipage}
\end{figure*}

Figure~\ref{fig:tasks_corr_wo_cpt} shows the Pearson correlation matrix of benchmark scores. The figure reveals that JEMHopQA, NIILC (QA about Japanese knowledge) and WMT20-en-ja (English-Japanese translation) are weakly correlated with other tasks. In addition, the figure shows strong correlations across languages in benchmarks of arithmetic reasoning (GSM8K vs. MGSM), academic subjects (MMLU vs. JMMLU), and code generation (HumanEval vs. JHumanEval). These findings are consistent with those identified with continually pre-trained LLMs in \S~\ref{sec:transferability}.

Figure~\ref{fig:pca_factor_score_wo_cpt} shows the factor loadings for each task benchmark.
The figure highlights four factors: a general ability factor with uniform scores on each benchmark (PC1); a Japanese ability factor with high scores on JEMHopQA, NIILC, and WMT20-en-ja (PC2); an ability factor for arithmetic reasoning and code generation with high scores on HumanEval, JHumanEval, MSGM, and GSM8K (PC3); and a residual factor that is difficult to interpret (PC4). These observations are consistent with those obtained with continually pre-trained LLMs in \S~\ref{sec:PCA}.

Lastly, we examined the relationship between the computational budget for English and PC1 (Figure~\ref{fig:budgetEN_f1_wo_cpt}) and the one between the computational budget for Japanese and PC2 (Figure~\ref{fig:budgetJA_f2_wo_cpt}).
Figure~\ref{fig:budgetEN_f1_wo_cpt} exhibits a strong positive correlation between PC1 (general ability) and computational budget for English ($\rho = 0.923$), and Figure~\ref{fig:budgetJA_f2_wo_cpt} indicates a moderate positive correlation between PC2 (Japanese ability) and computation budget for Japanese ($\rho = 0.779$). These relationships are the same as those confirmed with continually pre-trained LLMs in \S~\ref{sec:PCA_budget}.

In this way, we could confirm the findings observed in \S~\ref{sec:transferability} to \S~\ref{sec:PCA_budget} even with the LLMs built from scratch, which indicates the robustness of the findings against the construction methods of LLMs.

\subsection{Leave-One-Out Cross-Validation}
\begin{figure*}[t]
    \includegraphics[width=\linewidth]{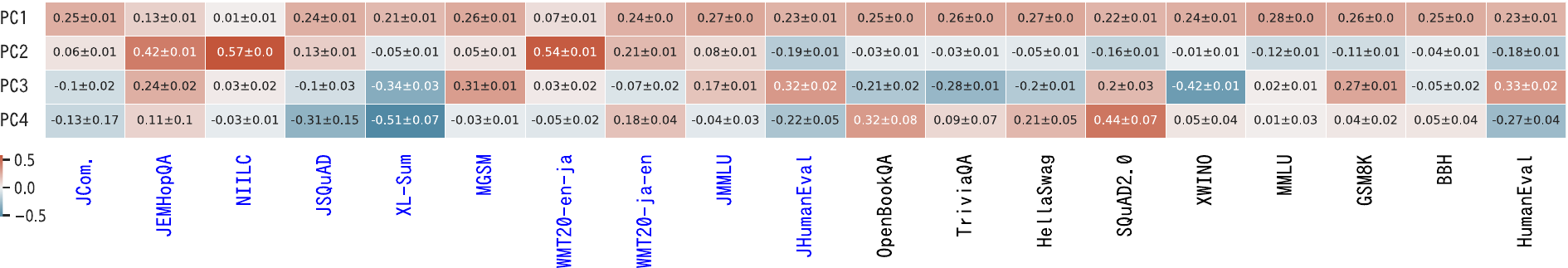}
    \caption{Leave-One-Out CV statistics: mean and standard deviations of the factor loadings ($n=35$, blue: Japanese benchmarks, black: English benchmarks).}
    \label{fig:loo}
\end{figure*}

We assessed the statistical error of factor loadings using leave-one-out cross-validation on the analyzed LLMs (see Figure \ref{fig:loo}) and confirmed that the standard deviations were small relative to the absolute values.

\end{document}